\documentclass[10pt,twocolumn,letterpaper]{article}

\usepackage[pagenumbers]{wacv} %
\usepackage{titletoc}
\usepackage{graphicx}
\usepackage{amsmath}
\usepackage{amssymb}
\usepackage{booktabs}

\usepackage[accsupp]{axessibility}

\usepackage{multicol}
\usepackage{microtype}
\usepackage{multirow}
\usepackage{wrapfig}

\usepackage{bm}
\usepackage{adjustbox}
\usepackage{array}
\usepackage{pifont}
\usepackage{float}
\usepackage{xfrac}
\usepackage{colortbl}
\usepackage[normalem]{ulem}
\usepackage{mathtools}
\usepackage{breqn}
\usepackage{stmaryrd}

\usepackage{caption}
\usepackage{subcaption}

\usepackage{comment}
\usepackage[dvipsnames,table]{xcolor}
\usepackage{soul}
\usepackage{cite}

\newcommand{\vtheta}{\boldsymbol{\theta}}

\newcommand{\vz}{\mathbf{z}}

\newcommand{\vy}{\mathbf{y}}
\newcommand{\vx}{\mathbf{x}}

\newcommand{\second}{\cellcolor{blue!10}}
\newcommand{\first}{\cellcolor{blue!30}}

\definecolor{dgreen}{rgb}{0.0, 0.5, 0.0}
\definecolor{better}{rgb}{0.19, 0.55, 0.91}
\definecolor{worse}{rgb}{0.82, 0.1, 0.26}
\newcommand{\cmark}{\textcolor{dgreen}{\ding{51}}}%
\newcommand{\xmark}{\textcolor{worse}{\ding{55}}}%

\usepackage[pagebackref,breaklinks,colorlinks]{hyperref}

\usepackage[capitalize]{cleveref}
\crefname{section}{Sec.}{Secs.}
\Crefname{section}{Section}{Sections}
\Crefname{table}{Table}{Tables}
\crefname{table}{Tab.}{Tabs.}

\def\wacvPaperID{1318} %
\def\confName{WACV}
\def\confYear{2024}

\usepackage{xcolor}

\begin{document}

\title{Learning to generate training datasets for robust semantic segmentation}
\author{\textbf{Marwane Hariat},\textsuperscript{\rm 1, *}  \textbf{Olivier Laurent},\textsuperscript{\rm 1, 2, *}  \textbf{Rémi Kazmierczak},\textsuperscript{\rm 1} \textbf{Shihao Zhang},\textsuperscript{\rm 3} \\
\textbf{Andrei Bursuc},\textsuperscript{\rm 4} \textbf{Angela Yao}\textsuperscript{\rm 3} \& \textbf{Gianni Franchi}\textsuperscript{\rm 1, $\dagger$} \\
 U2IS, ENSTA Paris, Institut Polytechnique de Paris,\textsuperscript{\rm 1} Université Paris-Saclay,\textsuperscript{\rm 2}  \\
National University of Singapore,\textsuperscript{\rm 3} valeo.ai\textsuperscript{\rm 4}
}
\twocolumn[{
\renewcommand\twocolumn[1][]{#1}%
\maketitle
\begin{center}
    \centering
    \captionsetup{type=figure}
    \captionsetup[subfigure]{labelformat=empty}
     \addtocounter{figure}{-1}
     \begin{subfigure}[b]{0.23\textwidth}
         \centering
         \includegraphics[width=\textwidth]{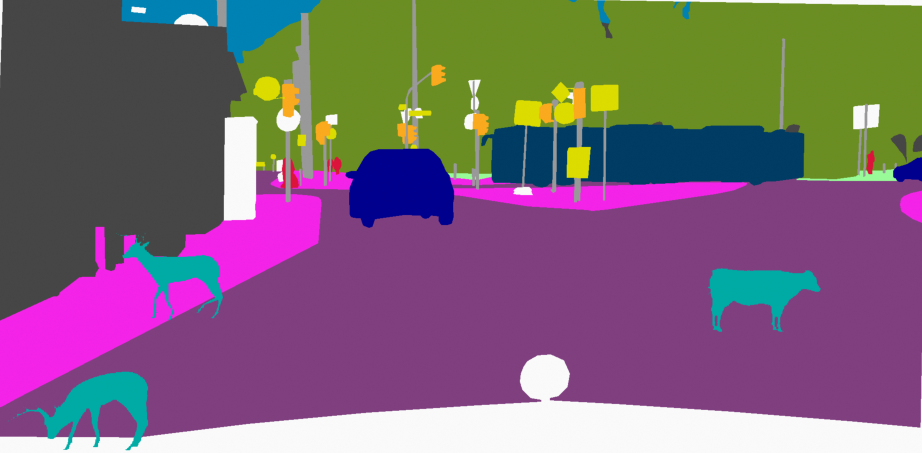}
         \label{figteaser:Label-anom}
     \end{subfigure}
     \hfill
     \begin{subfigure}[b]{0.22\textwidth}
         \centering
         \includegraphics[width=\textwidth]{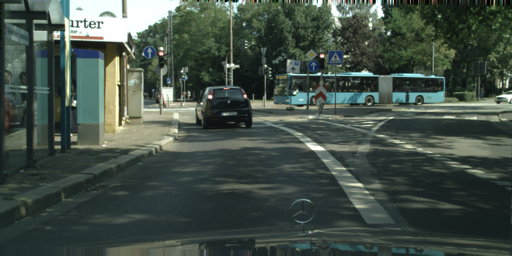}
         \label{figteaser:Orginalimage-anom}
     \end{subfigure}
     \hfill
     \begin{subfigure}[b]{0.22\textwidth}
         \centering
         \includegraphics[width=\textwidth]{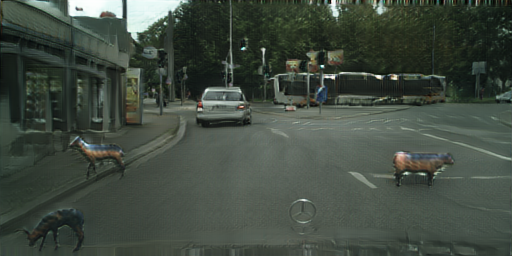}
         \label{figteaser:OasisGeneration-anom}
     \end{subfigure}
     \hfill
     \begin{subfigure}[b]{0.22\textwidth}
         \centering
         \includegraphics[width=\textwidth]{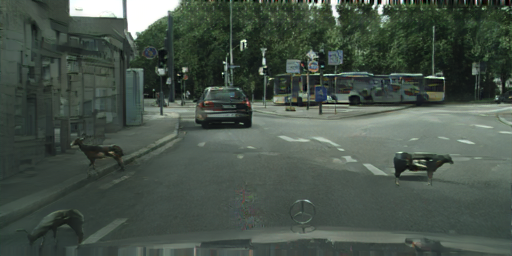}
         \label{figteaser:RobustaGeneration-anom}
     \end{subfigure}
     \hfill
         \begin{subfigure}[b]{0.23\textwidth}
         \centering
         \includegraphics[width=\textwidth]{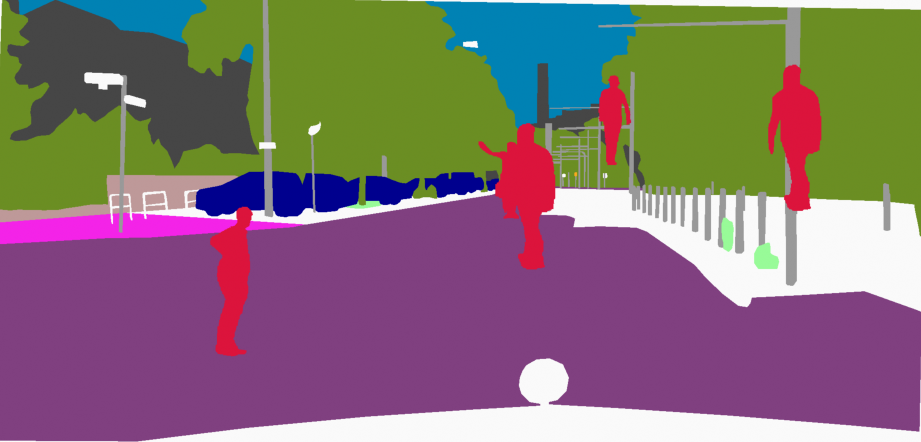}
         \caption{Segmentation maps}
         \label{figteaser:Label}
     \end{subfigure}
     \hfill
     \begin{subfigure}[b]{0.22\textwidth}
         \centering
         \includegraphics[width=\textwidth]{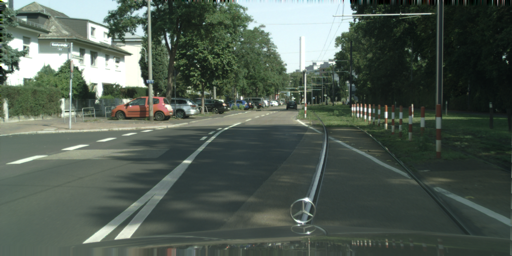}
         \caption{Original images}
         \label{figteaser:Orginalimage}
     \end{subfigure}
     \hfill
     \begin{subfigure}[b]{0.22\textwidth}
         \centering
         \includegraphics[width=\textwidth]{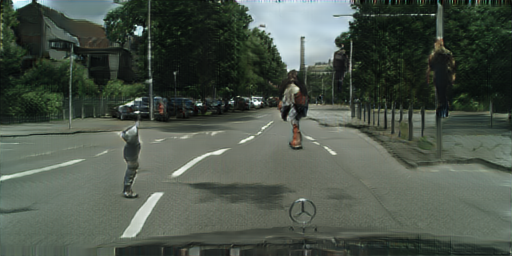}
         \caption{Oasis samples}
         \label{figteaser:OasisGeneration}
     \end{subfigure}
     \hfill
     \begin{subfigure}[b]{0.22\textwidth}
         \centering
         \includegraphics[width=\textwidth]{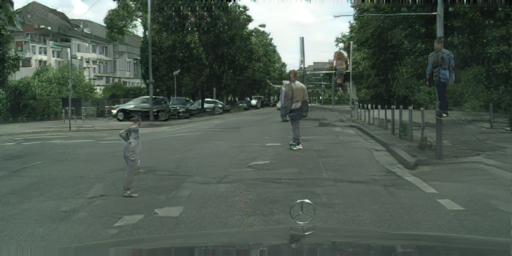}
         \caption{Robusta samples}
         \label{figteaser:RobustaGeneration}
     \end{subfigure}
    \captionof{figure}{\textbf{Illustration of image synthesis models under different perturbations.}
    Compared to previous work~\cite{sushko2021oasis}, our results express more natural textures and details, even under anomalies or shifts in the input distribution.
    }
    \label{fig:qualitative_robustness}
\end{center}}]
\def\thefootnote{*}
\begin{NoHyper}
\footnotetext{equal contribution, \quad \textsuperscript{$\dagger$} {\tt gianni.franchi@ensta-paris.fr}}
\end{NoHyper}
\def\thefootnote{\arabic{footnote}}

\begin{abstract}
Semantic segmentation methods have advanced significantly. Still, their robustness to real-world perturbations and object types not seen during training remains a challenge, particularly in safety-critical applications. We propose a novel approach to improve the robustness of semantic segmentation techniques by leveraging the synergy between label-to-image generators and image-to-label segmentation models. Specifically, we design \emph{Robusta}, a novel robust conditional generative adversarial network to generate realistic and plausible perturbed images that can be used to train reliable segmentation models. We conduct in-depth studies of the proposed generative model, assess the performance and robustness of the downstream segmentation network, and demonstrate that our approach can significantly enhance the robustness in the face of real-world perturbations, distribution shifts, and out-of-distribution samples. Our results suggest that this approach could be valuable in safety-critical applications, where the reliability of perception modules such as semantic segmentation is of utmost importance and comes with a limited computational budget in inference. We release our code at \href{https://github.com/ENSTA-U2IS-AI/robusta}{github.com/ENSTA-U2IS-AI/robusta}.

\end{abstract}

\vspace{-1.35cm}
\section{Introduction}
Semantic segmentation is an essential perception task that is commonly used in safety-critical applications such as autonomous driving~\cite{levinson2011towards, mcallister2017concrete} or medical imaging~\cite{maag2020time, oktay2018attention}. To fulfill safety requirements~\cite{lavin2022technology}, it is crucial to not only produce accurate segments but also make reliable predictions in the face of perturbations~\cite{hendrycks2019robustness}, distribution shifts~\cite{tremblay2021rain, hu2019depth, sakaridis2018semantic}, uncommon situations~\cite{zendel2018wilddash} and out-of-distribution (OOD) objects~\cite{hendrycks2019anomalyseg,chan2021segmentmeifyoucan}. 
Modern Deep Neural Networks (DNNs) achieve impressive performance in segmentation tasks~\cite{farabet2012learning, ren2015faster, he2017mask, badrinarayanan2017segnet, orvsic2021efficient}; however, they struggle to generalize to data samples not seen during training, e.g., image corruptions~\cite{pezzementi2018putting,hendrycks2019robustness}, adversarial attacks~\cite{szegedy2013intriguing,eykholt2018robust}, or change of style~\cite{geirhos2018imagenet}. In the face of such events, they often produce overconfident probability estimates~\cite{nguyen2015deep,guo2017calibration,hein2019relu} even when they are wrong, which can impede the detection of failure modes and thus their adoption in the industry.

Technically, modern high-capacity DNNs trained with uncertainty and calibration awareness could effectively learn to deal with such situations if they were seen during training. However, such samples from the long tail are extremely numerous, difficult, and expensive to acquire in the real world. In the absence of such data, new data augmentation techniques have been devised towards mitigating this problem: generating perturbations on the input data~\cite{zhang2018mixup,yun2019cutmix}, the latent space~\cite{du2022vos}, or the output space~\cite{lee2018training}. They may slightly relieve this problem by increasing the size of the training set with richer samples. Still, they may not always yield realistic images and are often architecture-dependent or unsuitable for semantic segmentation. 
Meanwhile, generative models are making steady progress towards the ambitious goal of generating abundant and high-quality data for training data-hungry DNNs~\cite{brock2019large, karras2021stylegan3, ramesh2021zero,rombach2022high}. Despite the realism reflected by excellent FID scores~\cite{heusel2017gans}, training on such synthetic images is non-trivial and less effective than with real images~\cite{besnier2020dataset, ravuri2019classification, shmelkov2018good}. Very recent methods based on the latest generative models~\cite{sariyildiz2022fake,he2023synthetic} are starting to show encouraging results for image classification. However, such approaches have not %
shown yet that they can improve the performance and robustness of semantic segmentation models dealing with high-resolution images and complex scenes.

In this paper, we pave the way for improving the robustness of semantic segmentation models against input perturbations as well as their ability to detect outliers, i.e., objects with no associated labels in the train %
set. To this end, we propose leveraging the symmetry between label-to-image conditional generative adversarial networks (cGANs)~\cite{goodfellow2014generative,isola2017image} and image-to-label models to train robust %
segmentation networks~\cite{lis2019detecting,di2021synboost}. We argue that it is possible to use label-to-image %
cGANs at training time to enrich %
datasets with perturbations and outliers %
and improve %
segmentation models. 
The generated data must be of sufficient quality and diversity to enable the classifier - trained on it - to generalize to new and unseen real-world situations and to prevent overfitting to the specific anomalies of the training set. 

A significant difficulty we face is that cGANs are not robust to variations in the input labels. They may generate images with artifacts when given label maps with anomalies, hindering their utility for semantic segmentation training. To address this challenge, we advance \emph{Robusta}, a novel cGAN architecture that leverages attention layers~\cite{vaswani2017attention} and sub-networks~\cite{wang2018high}. Robusta 
can produce high-quality images even from label maps with corruptions and anomalies. We 
exploit Robusta synthetic images 
to construct a dataset for training an observer network~\cite{besnier2021triggering} to detect anomalies and failures of the segmentation model.

\vspace{0.2cm}
\noindent\textbf{Contributions.} 
\textbf{(1)} We propose a new strategy to improve the robustness and OOD detection performance of semantic segmentation models by leveraging the symmetry of label-to-image cGANs and image-to-label segmentation networks: we use cGAN images with outliers to train a robust segmentation model; \textbf{(2)} We design the first framework to evaluate the robustness of label-to-image cGANs against perturbations and uncommon inputs, and we use it to investigate the robustness of multiple cGAN methods against three new dataset generation techniques; \textbf{(3)} We propose \emph{Robusta}, a new cascaded cGAN with improved robustness compared to state-of-the-art cGANs on the proposed framework. Our approach is expected to enhance the reliability and safety of autonomous driving systems by enabling more accurate OOD and failure detection in real-world situations.

\section{Related Work}

\vspace{0.2cm}
\noindent\textbf{Conditional GANs for label-to-image translation:} Various label-to-image translation techniques have been proposed~\cite{pang2021image} with two main categories: paired strategies~\cite{isola2017image, spadepark2019semantic,pang2021image,wang2022pretraining,zhou2021cocosnet}, where images and labels are aligned, and unpaired strategies~\cite{zhu2017unpaired,jiang2020tsit}%
. This work focuses on paired label-to-image translation models that synthesize RGB images from semantic inputs. Most approaches in this area are based on cGANs~\cite{isola2017image, spadepark2019semantic}. 

Pix2Pix~\cite{isola2017image} is a strong baseline and was one of the first cGAN-based techniques applied to label-to-image translation, but it often fails to generate high-quality images using a single GAN. Pix2PixHD~\cite{wang2018high}
upgrades the quality of the generated images by incorporating enhanced multi-scale generators and discriminators.

SPADE (SPatially ADaptive DEnormalization)~\cite{spadepark2019semantic} advances spatially-adaptive normalization layers that leverage semantic label maps to modulate the activations in the normalization layers, resulting in significant improvements for datasets with strong global redundancy. Rather, DP-GAN~\cite{schonfeld_sushko_iclr2021} shares the spatially adaptive normalization parameters, learned at each scale of the down-sampling part, with the up-sampling layers.
More recently, OASIS~\cite{sushko2021oasis} incorporates an adversarial component and pixel-level discriminators, and PITI~\cite{wang2022pretraining} pre-trains a diffusion model on a large dataset of various images while leveraging the latent space for the downstream tasks.

\vspace{0.2cm}
\noindent\textbf{Anomaly detection for semantic segmentation:} 
Anomaly segmentation is challenging as it requires precise anomaly localization and can be tackled by cGAN-based methods. Various cGAN methods~\cite{lis2019detecting, xia2020synthesize,vojir2021road,di2021synboost} have been proposed to detect anomalies. These methods utilize cGANs at test time, leading to significant runtime costs~\cite{tian2022pixel}. In contrast, our approach only uses cGANs in the offline stage to generate synthetic images for training an efficient anomaly-aware segmentation network, avoiding the runtime costs associated with cGAN-based approaches.

\begin{figure*}[ht!]
\centering
\begin{center}
\includegraphics[width=\linewidth]{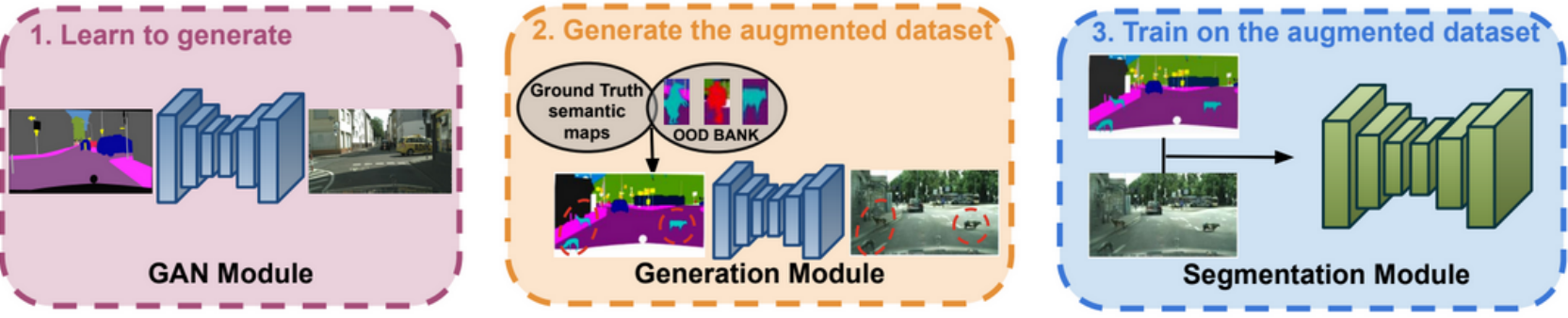}
\end{center}
\caption{\textbf{Illustration of the pipeline.} The pipeline diagram depicts three steps. Firstly, we train the new Robusta model. Secondly, we utilize Robusta to create a diverse, high-quality dataset that includes various objects of interest. Finally, we train a segmentation model on this augmented dataset.}
\label{fig:pipeline0}
\end{figure*}

Other techniques employ uncertainty quantification methods such as ensembles~\cite{lakshminarayanan2017simple}, their efficient derivatives~\cite{huang2017snapshot,garipov2018loss,wen2020batchensemble,franchi2020encoding,havasi2020training,laurent2023packed}, and Monte-Carlo Dropout~\cite{gal2016dropout,gal2017concrete,mukhoti2018evaluating}. Some approaches, such as DNP~\cite{galesso2022far}, compute distances between feature representations produced by the segmentation encoder network~\cite{reiss2021panda}  using nearest-neighbors. Others, like PEBAL~\cite{tian2022pixel}, utilize abstention learning ~\cite{liu2019deep}  at the pixel level through energy-based models~\cite{lecun2007energy}.

Some methods~\cite{bevandic2019simultaneous,grcic2022densehybrid,chan2021entropy} involve using auxiliary data in the wake of Outlier Exposure~\cite{hendrycks2018deep} for classification. Going further, Obsnet~\cite{besnier2021triggering} generates failures with local adversarial attacks and NflowJS~\cite{grcic2020dense, grcic2021dense}, generate a negative dataset using normalizing flows~\cite{rippel2013high, dinh2014nice, papamakarios2021normalizing}. In contrast, our method employs cGANs to generate additional data with non-typical content and scene layouts, %
and generates outliers as a proxy for out-of-distribution. This generated data is used downstream %
to train robust segmentation models.

\vspace{0.2cm}
\noindent\textbf{Dataset synthesis for semantic segmentation:} The lack or limited amount of anomalous data in real-world datasets motivates the synthesis of datasets for semantic segmentation. StreetHazards~\cite{hendrycks2019scaling} and MUAD~\cite{franchi2022muad} are recent works that propose solving this problem with fully synthetic datasets generated with computer graphics simulators.

To increase the volume of data of a given dataset and test the robustness of segmentation models, some researchers propose using GANs to modify images with challenging conditions of interest. For instance, Rainy~\cite{tremblay2021rain} and Foggy Cityscapes~\cite{sakaridis2018semantic} respectively add rain and fog to Cityscapes' urban scenes, and CoMoGAN~\cite{pizzati2021comogan} allows %
changing the picture's time of the day with continuous transformations.

An alternative to these methods consists of directly generating full scene layouts. SB-GAN~\cite{azadi2019semantic} and PGAN-CGAN~\cite{howe2019conditional} are recent methods proposed for this task, generating semantic label maps and converting them into images using classical cGANs. SemanticPalette~\cite{le2021semantic} improves upon these methods with a pipeline controlling the relative importance of the different image classes.

\section{Problem setup and method overview}\label{sec:setup}

Our objective is to improve the ability of semantic segmentation models to handle unexpected and uncommon situations and objects that arise due to aleatoric and epistemic uncertainty sources.
For the scope of this paper, we define a robust semantic segmentation model as one that can produce accurate predictions even when confronted with uncommon samples coming from a distribution with strong %
uncertainties, shifted from the original training distribution of the model. To consolidate the robustness of this segmentation model, we introduce a three-step pipeline, which we outline below and in Figure~\ref{fig:pipeline0}.

\textbf{Step 1:}  We train a new cGAN called Robusta to generate additional data from the original segmentation training set (see section \ref{ssec:Robusta}). \textbf{Step 2:}  We leverage Robusta to create a new high-quality dataset by conditioning it on label maps with different %
objects, such as road signs %
placed in unconventional locations, aiming to expand the diversity of the training set (see section \ref{ssec:eval_protocol_robust}). \textbf{Step 3:}  We boost the generalization ability of the segmentation model by training it on this new augmented dataset (see section \ref{sec:seg}).

To the best of our knowledge, we are the first to explore the robustness of %
generative models in the context of semantic segmentation for autonomous driving scenes. %
To this extent, we introduce a novel architecture that enhances the reliability of generated images. Our primary goal is, indeed, to develop a 
robust label-to-image generative model capable of producing realistic corner-case images to improve the robustness of segmentation networks. To assess the effectiveness of these models and our pipeline (see Figure \ref{fig:pipeline0}), we employ three benchmarks: first, we quantify the quality of the generated images under non-perturbed conditions (purple block) in Appendix \textbf{C}; then we measure the robustness of the generative model with perturbed label-maps (yellow block) in sections \ref{ssec:eval_protocol_robust} and \ref{ssec:exp2RobustnessGAN}; finally, we evaluate the efficiency of synthetic images in enhancing the robustness of semantic segmentation (blue block) in section \ref{ssec:exp3Robustnessaleatoric}.

\begin{figure*}[t]
\centering
\begin{center}
\includegraphics[width=0.8\linewidth]{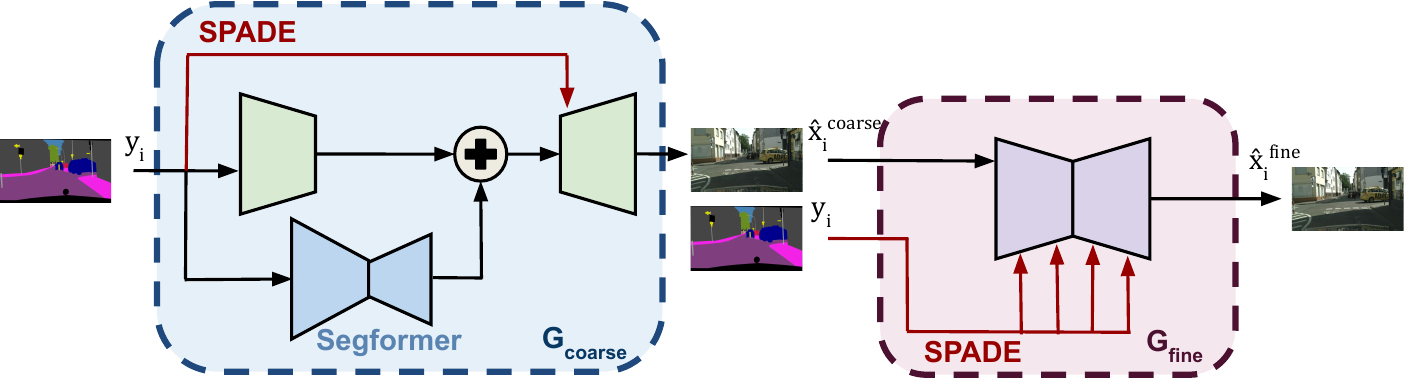}
\end{center}
\caption{\textbf{Illustration of Robusta's generation process.} First, we train the networks $G_{\text{coarse}}$, producing low-resolution images. We add another generator, $G_{\text{fine}}$, to improve the image quality from the output of $G_{\text{coarse}}$. The \textbf{+} operation corresponds to a concatenation.}
\label{fig:robusta}
\end{figure*}

\section{Enhancing data generation to robustify segmentation}

The efficient training and robustification of the semantic segmentation network rely on generating high-quality data (see Appendix \textbf{F.1} for more details). This requires generating both in-distribution data with perturbations and different textures and high-quality out-of-distribution data. As outlined in Section~\ref{sec:setup}, enhancing the generating models' robustness is essential. Specifically, the quality of the output must remain satisfactory even when given inputs outside %
of its training distribution (corrupted label maps). 
In section \ref{ssec:Robusta}, we propose a novel cGAN-cascade called \emph{Robusta} to improve the robustness of label-to-image translation. We test the usefulness of our method thanks to the first \textbf{protocol to evaluate the robustness of generative processes} that we detail in section \ref{ssec:eval_protocol_robust}. 
To enhance the robustness of the cGAN, we have integrated attention layers~\cite{vaswani2017attention} and decomposed the GAN into two sub-GANs.

\subsection{Robusta: a new cGAN cascade for improved robustness}\label{ssec:Robusta}

\subsubsection{Architecture}
We propose \emph{Robusta} a novel architecture based on cGANs %
designed to address the artifacts presented in Figure \ref{fig:qualitative_robustness} that arise when generating outlying %
label maps in the context of label-to-image translation. To overcome these artifacts, we decompose our translation cascade into two generators $G_{\text{coarse}}$ and $G_{\text{fine}}$, as illustrated in Figure \ref{fig:robusta}, that are trained sequentially.
We design $G_{\text{coarse}}$ for handling artifacts related to label-to-image translation and $G_{\text{fine}}$ artifacts related to output image quality.

The architecture of $G_{\text{coarse}}$ is based on Pix2PixHD~\cite{wang2018high} %
and includes two encoder-decoders (in blue and green in Figure \ref{fig:robusta}). These encoder-decoders have the same input dimensions, but the output of the blue module has the size of the latent space of the green module. For better robustness and performance, we choose Segformer~\cite{xie2021segformer} for the blue module as encoder-decoder, which has proven effective in semantic-segmentation tasks. Segformers %
are based on ViTs~\cite{dosovitskiy2020image} and encode inputs in feature maps, which promotes contextual information learning~\cite{li2022contextual,ghiasi2022vision}. Moreover, ViTs are more robust learners~\cite{bhojanapalli2021understanding, bai2021transformers, benz2021adversarial}.
We think %
that the favorable Transformer's robustness also extends to image generation, %
and support this hypothesis with the ablation studies conducted in Appendix \textbf{E.1}.
We argue that this approach is particularly relevant for tackling translation artifacts. Additionally, we incorporate a SPADE~\cite{spadepark2019semantic} layer after the concatenation of the output of the Segformer (in blue) and the encoding of the main generator (in green) to re-inject lost spatial semantic information before decoding. 

To further improve the quality of the final image, we add a generator, $G_{\text{fine}}$, which takes as input the images provided by $G_{\text{coarse}}$ and generates high-quality outputs. 
In $G_{\text{fine}}$, we substitute all batch normalizations~\cite{ioffe2015batch} %
with spatially-adaptive normalization (SPADE)~\cite{spadepark2019semantic} layers %
to accurately incorporate the information present in %
semantic label maps. 

To obtain a better image quality with higher-frequency textures~\cite{park2022vision} (see Appendix~\textbf{E.1.3} for more details), we use U-Net~\cite{ronneberger2015u,saharia2022photorealistic}, a residual convolutional neural network~\cite{he2016deep}. $G_{\text{fine}}$ is trained separately from $G_{\text{coarse}}$ using the loss \eqref{eq:LG2}, $G_{\text{coarse}}$'s weights being frozen at this time.

Appendix~\textbf{A} details the links between Pix2Pix, Pix2PixHD, SPADE, and Robusta, and Appendix~\textbf{E.1} provides ablation studies on the architecture.

\begin{figure*}[t]
\centering
\begin{tabular}{llll}
\includegraphics[width=0.23\linewidth]{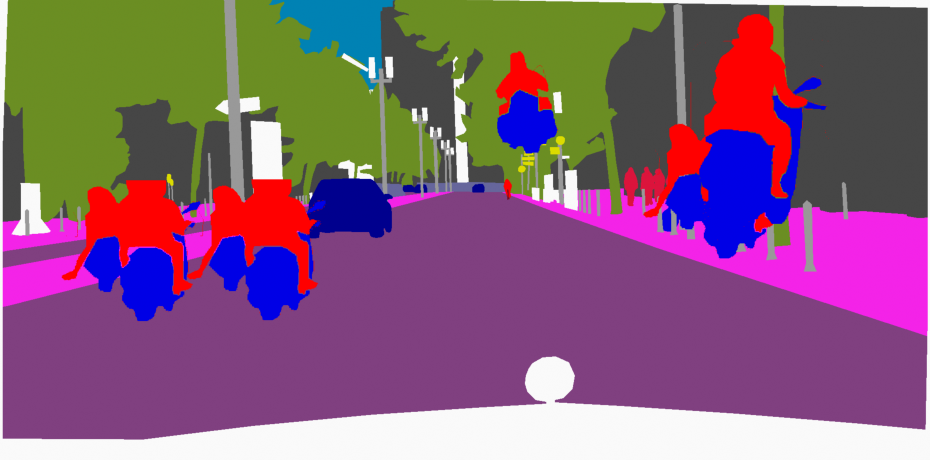} & \includegraphics[width=0.23\linewidth]{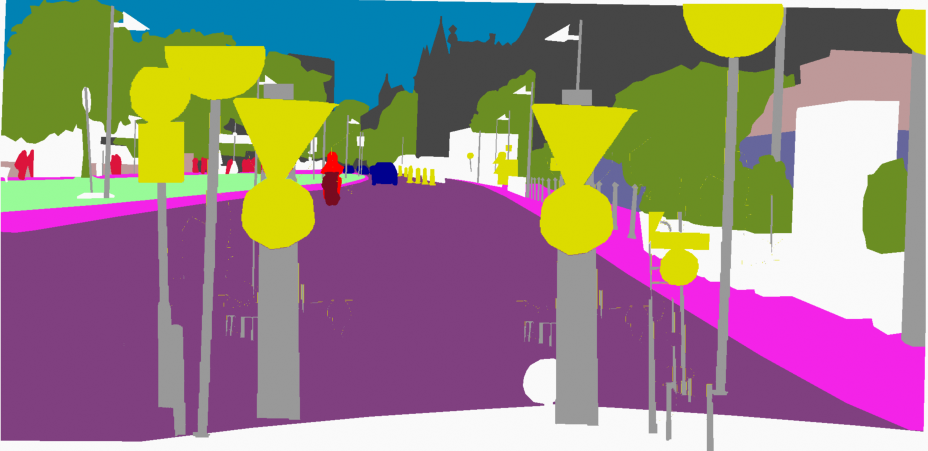} & 
\includegraphics[width=0.23\linewidth]{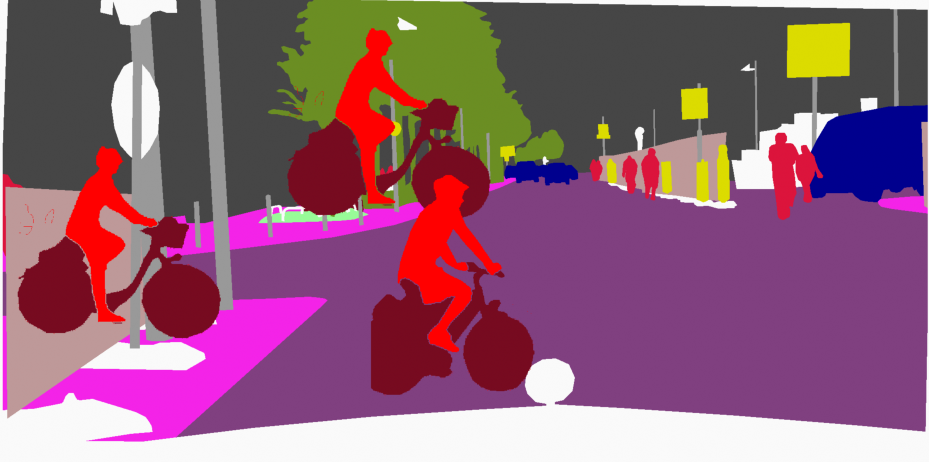}
\includegraphics[width=0.23\linewidth]{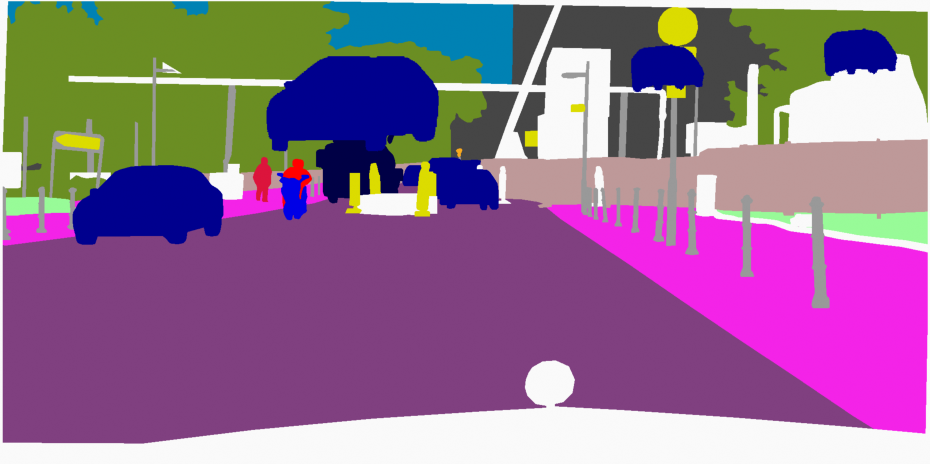} \\
\includegraphics[width=0.23\linewidth]{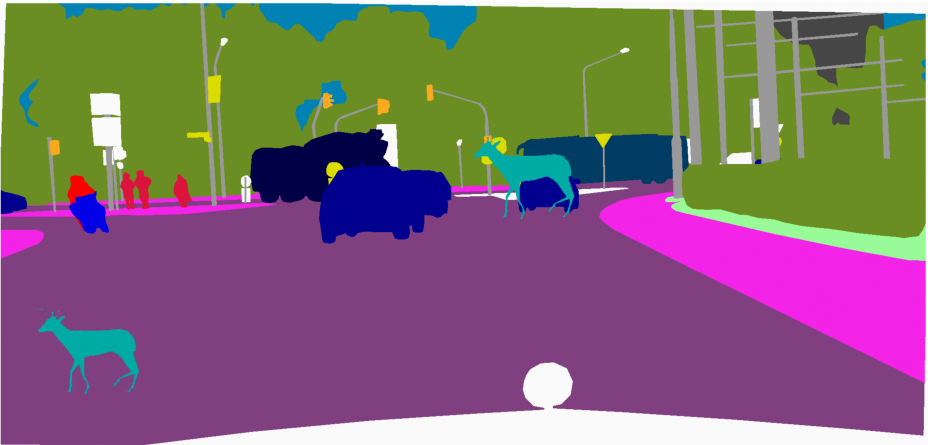} & \includegraphics[width=0.23\linewidth]{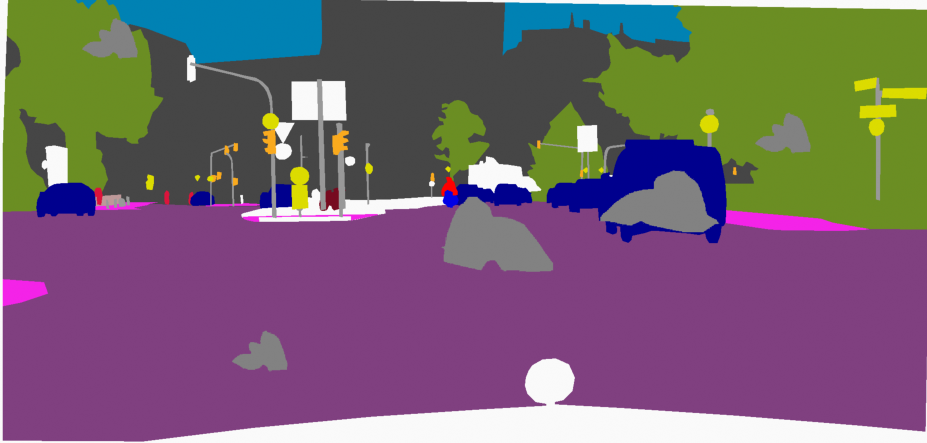} & 
\includegraphics[width=0.23\linewidth]{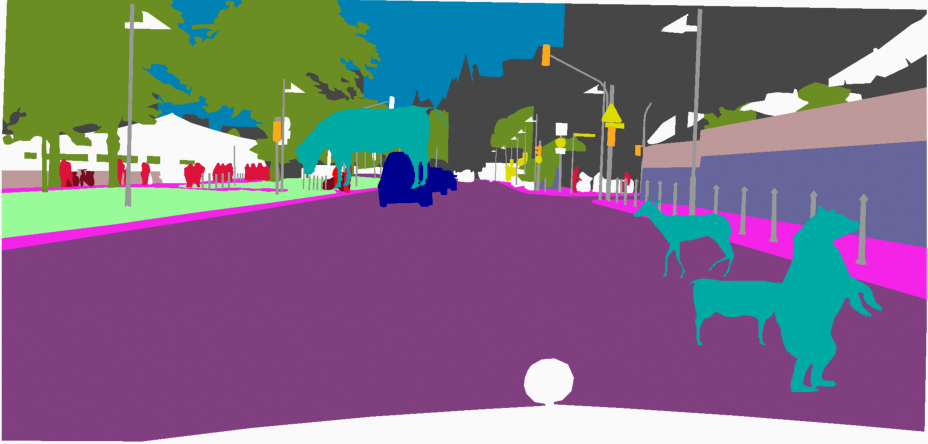}
\includegraphics[width=0.23\linewidth]{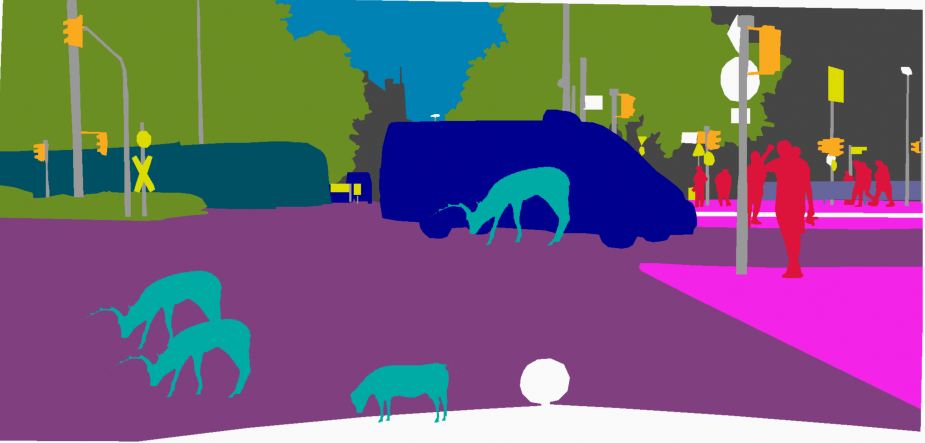} 
\end{tabular}
\caption{\textbf{Perturbed label maps.} Different label maps from Corrupted-Cityscapes (top) and Outlier-Cityscapes (bottom). 
}
\label{fig:monodepth}
\end{figure*}

\subsubsection{Training}
\paragraph{Standard procedure.}
Let $F_{\vtheta}(\vx)$ denote the application of a DNN $F$ of weights $\vtheta$ to an input image $\vx$, $D_{\vtheta}(\cdot)$ %
a discriminator %
and $G_{\vtheta}(\cdot)$ %
a generator%
. The weights of the discriminator and the generator are denoted by $\vtheta_D$ and $\vtheta_G$ respectively and are omitted when obvious%
.
Label-to-image models consist of %
cGANs trained with the following losses:

\noindent(a) A GAN loss:
\begin{multline}\label{eq:GANloss}
\mathcal{L}_{\text{cGAN}}^{\text{BCE}}(\vtheta_G, \vtheta_D) =\mathbb{E}_{\vx}[\log D(\vx \mid \vy)]+ \\ \mathbb{E}_{\vz}[\log (1-D(G(\vz \mid \vy)))],
\end{multline}
\noindent
where $\vy$ is the conditional information. In label-to-image translation, $\vy$ is the input label, $\vx$ is the target RGB image, and $\vz$ is the sampled latent variable.\\
(b) An L1 loss between the output of the generator and the target images, denoted $\mathcal{L}_{L1}(\vtheta_G)$.\\
(c) A feature-matching loss $\mathcal{L}_{\mathrm{FM}}\left(\vtheta_G, \vtheta_{D}\right)$
measuring the L1 norm between the feature maps of the real and generated images extracted from the layers of the discriminator D.\\
(d) A perceptual loss %
$\mathcal{L}_{\mathrm{VGG}}$ %
measuring the L1 norm between the feature maps of the real and generated images extracted from the layers of a VGG network~\cite{simonyan2014very} pre-trained on ImageNet~\cite{deng2009imagenet}. \\

\vspace{-0.5cm}
\paragraph{Training Robusta.}
We train Robusta in two steps beginning with the generator $G_{\text{coarse}}$. To do so, we optimize the loss $\mathcal{L}_{\text{coarse}}$ on a dataset made of the original label maps as inputs and original images as targets. We minimize the linear combination of losses \eqref{eq:LG1} of Wang et al.~\cite{wang2018high}.%
\begin{equation}\label{eq:LG1}
    \mathcal{L}_{\text{coarse}} = \mathcal{L}_{\text{cGAN}}^{\text{BCE}} + \lambda_{\text{L1}} \mathcal{L}_{\text{L1}} + \lambda_{\textit{FM}}\mathcal{L}_{\text{FM}} + \lambda_{\text{VGG}}\mathcal{L}_{\text{VGG}}. 
\end{equation}
When the training of $G_{\text{coarse}}$ is complete, the synthesized images are used to feed and train the second generator $G_{\text{fine}}$, which does not use $\mathcal{L}_{L1}$ and is based on $\mathcal{L}_{\text{cGAN}}^{\text{MSE}}$ instead of $\mathcal{L}_{\text{cGAN}}^{\text{BCE}}$ as proposed for SRGAN~\cite{ledig2017photo}. This corresponds to:%
\begin{equation}\label{eq:LG2}
    \mathcal{L}_{\text{fine}} = \mathcal{L}_{\text{cGAN}}^{\text{MSE}} +
    \lambda_{\text{FM}}\mathcal{L}_{\text{FM}} + \lambda_{\text{VGG}}\mathcal{L}_{\text{VGG}}. 
\end{equation}
Please refer to the Appendix \textbf{A} for details on the losses and the definition $\mathcal{L}_{\text{cGAN}}^{\text{MSE}}$, and to Appendix \textbf{B} for training details. Furthermore, Appendix \textbf{E.2} includes ablation studies on Robusta's training, and Appendix \textbf{F.3} details Robusta's training time and computational cost.

\subsection{Robustness assessment protocol for label-to-image translation}\label{ssec:eval_protocol_robust}

We present the first protocol to evaluate the robustness of label-to-image generators quantitatively. This type of network is susceptible to perturbations in input data, prompting us to design three novel methodologies to gauge the efficacy of label-to-image generators derived from semantic segmentation datasets. These methodologies involve modifying the original dataset to quantify the robustness of generators against diverse sources of corruption.

When working on Cityscapes~\cite{cordts2016cityscapes} (CS), we can process the images of the validation set and produce three versions: \emph{Corrupted-CS}, \emph{Outlier-CS}, and \emph{Morphological-CS}. Note that the following processes naturally extend any other semantic segmentation dataset. 

\emph{Corrupted-CS} randomly adds Cityscapes objects, such as road signs and cyclists, into the label maps. A robust label-to-image network should be able to properly render the additional objects that are included in its training set.

\emph{Outlier-CS} adds objects that do not belong to the distribution of the original dataset. As Cityscapes is composed of urban scenes, we add shapes of bears, deer, cows, and rocks. With this dataset, we evaluate the capacity of the network to generate renderings from unknown shapes. %

Last, \emph{Morphological-CS} corrupts the segmentation maps by applying morphological operators on the labels. We detail the morphological operations in Appendix~\textbf{D}. As for CIFAR-C~\cite{hendrycks2019robustness} and Cityscapes-C~\cite{michaelis2019benchmarking}, we study the robustness of the network to varying intensities of perturbations of the input images. In our case, this corresponds to increasing the size of the structuring element of the morphological dilation.

In these three variations, we expect a robust label-to-image translator to produce reliable and high-quality outputs even when perturbed by modified label maps. Hence, we measure the quality of the generated images using the Fréchet Inception Distance (FID). We also assess the semantic segmentation performance of a pre-trained model on the original dataset but inferring on the modified version, with the mean Intersection over Union (mIoU). This value should remain high if the rendering conveys the semantic meaning correctly. Corrupted label maps are available in Figure \ref{fig:monodepth}.

Our method generates artifacts and out-of-domain object maps, but we claim that they are relevant to enrich training data and %
robustify training. We view these perturbations %
of the true label maps distributions as targeting the critical long-tail anomalies. For instance, flying cars could be related to car crashes in real-world images. In our training pipeline (on Figure \ref{fig:pipeline0}), achieving good quality on these datasets is %
crucial for the subsequent training of the segmentation methods. Therefore, we use these datasets as a benchmark for evaluating the robustness of our generator models.

The generating process of the conditional GAN is thoroughly explained in Appendix~\textbf{A}. Additionally, section \ref{sec:seg} provides a detailed account of the techniques used for generating images of the \emph{Outlier-CS} and \emph{Corrupted-CS} datasets.

\section{New training sets for robust segmentation}\label{sec:seg} 

Thanks to Robusta, it is now possible to generate high-quality data with outliers and perturbations to extend existing segmentation datasets while limiting artifacts. The primary goal is to harness this generated data and refine the training process to obtain more resilient segmentation models.

\begin{table*}[t]
\setlength{\abovecaptionskip}{0.cm}
\begin{center}
\resizebox{0.95\textwidth}{!}{
	\begin{tabular}{@{}ccccccccccccc<{\kern-\tabcolsep}}
    \toprule
    \multirow{2}{*}{Method }  & \multicolumn{2}{c}{ Corr.-CS } & \multicolumn{2}{c}{ Corr.-ADE20K } & \multicolumn{2}{c}{  Outlier-CS } & \multicolumn{2}{c}{  Outlier-ADE20K }& \multicolumn{2}{c}{ Morph.-CS}& \multicolumn{2}{c}{ Morph.-ADE20K }\tabularnewline
    &    FID $\downarrow$  &  mIoU $\uparrow$ &    FID $\downarrow$  &  mIoU $\uparrow$ & FID $\downarrow$  &  mIoU $\uparrow$ &  
     FID $\downarrow$  &  mIoU $\uparrow$ & 
     FID $\downarrow$  &  mIoU $\uparrow$  & 
     FID $\downarrow$  &  mIoU $\uparrow$  \tabularnewline
    \midrule

    { SPADE}    & 77.4  & \second42.7  & 46.3  & 33.2 & 91.8  & 59.2 & 52.0  & 37.7  &  61.8  & 59.6  & 40.1  &  38.5 
    \tabularnewline

    OASIS	  & 76.9  & 41.8  & 45.9  & \second44.1 & 90.0  & 62.0  & 67.3  & \second47.1    & \second51.9  & \second62.9 &  39.0  &  \second48.7 \tabularnewline
    \midrule

	Robusta $G_{\text{coarse}}$	  & \second 76.5  & 42.0 &\second 40.8 & 42.7 & \second84.7  & 60.4  & \second45.1  & 45.5  & 53.6  & 58.1 &  \second35.8  &  46.1 \tabularnewline

	Robusta $(G_{\text{coarse}}, G_{\text{fine}})$	  & \first \textbf{75.3}  & \first\textbf{43.3}  & \first\textbf{38.1}  & \first\textbf{44.6}  & \first\textbf{79.5}  & \first\textbf{62.8}   & \first\textbf{43.2}  &\first \textbf{47.3} & \first\textbf{50.6}  & \first\textbf{63.4}  &  \first\textbf{33.9}  &  \first\textbf{48.8} \tabularnewline
 
 \bottomrule

\end{tabular}

 }
\end{center}
\vspace{-2mm}
\caption{\textbf{Comparative results for the robustness tasks} presented in \ref{ssec:eval_protocol_robust}. Corr. stands for corrupted and morph. for morphological.}
\label{table:res_robustness_GAN}
\vspace{-2mm}
\end{table*}

To train segmentation %
networks, we begin with a dataset of training examples $\mathcal{D}=\{\vx_i, \vy_i\}_{i=1}^{|\mathcal{D}|}$ %
of $|\mathcal{D}|$ pairs of images $\vx_i\in\mathbb{R}^{C\times H\times W}$ and label maps $\vy_i\in \llbracket 0, N_C \llbracket^{H\times W}$ modeled as a realization of a joint distribution $\mathcal{P}{(X,Y)}$. We denote $C$, $H$, and $W$ as the number of channels, the height, and the width of the image, respectively, and $N_C$ is the number of classes in the dataset.

As described in \ref{sec:setup}, the generation pipeline $G$ detailed in Figure \ref{fig:robusta} is designed to generate realistic images that correspond to the input label maps $\vy_i$, producing a $\hat{\vx}_i$ image. In this section, %
we aim to use $G$ to generate a synthetic dataset as depicted in Figure \ref{fig:pipeline0}. We use Cityscapes as our original training set for illustration. In section~\ref{ssec:eval_protocol_robust}, we previously introduced \emph{Corrupted-CS} and \emph{Outlier-CS}. Here, we provide more details about their use and definition. In the following subsections, we clarify how $G$ %
generates these two sets.

\vspace{-0.2cm}
\paragraph{Improving generalization with \emph{Corrupted-CS}:}

To enhance robustness, we generate \emph{Corrupted-CS} using Robusta.
First, we gather multi-dimensional masks of objects belonging to a particular set of classes $\mathcal{C}$. These classes may consist of various combinations, such as traffic signs and poles, traffic lights and poles, motorcycles and riders, and bicycles and riders. The collected masks are then sorted into sets $\mathcal{D}^{\text{OI}} \triangleq \{y^{\text{OI}}_i\}$ , where each $y^{\text{OI}}_i$ is an element of the interval $\llbracket 0, N_C \llbracket^{H\times W}$. All pixels of $y^{\text{OI}}_i$ except for the object's pixel are set to zero, whereas the object's pixel is assigned the label's object value.
Then, we expand $\mathcal{D}^{\text{OI}}$ by applying translations to its masks. Next, we use a mixing function to combine the label maps of the original dataset $\mathcal{D}$, creating a new label $\vy^{\text{mix}}_i$ for each element $i$ in the range $[1,|\mathcal{D}|]$. The definition of this new label is as follows:
\begin{equation}
\label{eq:mix0}
\vy^{\text{mix}}_i = \left[\vy^{\text{OI}}_j=0\right]*\vy_i + \vy^{\text{OI}}_j \mbox{, with } j \in [1,|\mathcal{D}^{\text{OI}}|]
\end{equation}
This results in a new set of label maps denoted as $\mathcal{D}^{\text{mix}}_y$.
Finally, we combine $G(\mathcal{D}^{\text{mix}}_y)$ for the images and $\mathcal{D}^{\text{mix}}_y$ for the labels to create the final dataset $\mathcal{D}^{\text{mix}} = \{G(\vy^{\text{mix}}_i), \vy^{\text{mix}}_i\}_{i=1}^{|\mathcal{D}|}$. 

To improve the generalization of the semantic segmentation model, we enrich the original dataset $\mathcal{D}$ with $\mathcal{D}^{\text{mix}}$, and we keep the same training procedure. 

\vspace{-0.2cm}
\paragraph{Improving outlier detection with \emph{Outlier-CS}:}

To enhance the detection of outliers, we craft Outlier-CS. We apply the same mixing strategy as before, %
except that we no longer extract known objects to construct $\mathcal{D}^{OI}$. Instead, we choose a set of outlier objects from Cityscapes, such as bears, cows, and rocks. The shape of these instances is derived from the MUAD dataset~\cite{franchi2022muad}. Since outliers do not possess labels %
corresponding to the CS label space, we value them as "humans" using the aforementioned outlier shapes.

To improve out-of-distribution detection, we can also use the outlier labels to train the networks directly with a binary cross-entropy (BCE) loss~\cite{cox1958regression} and refer to this method as "Robusta + BCE". In the following, we test the efficiency of adding this dataset for Obsnet-like~\cite{besnier2021triggering} architectures on OOD detection tasks.

Contrary to previous work by Ghiasi et al.~\cite{ghiasi2021simple}, which requires finely labeled object instances, Robusta can generate instances even from unknown shapes, eliminating the need for expensive fine-grained annotations of OOD instances. These annotations can be more expensive than segmentation labels, especially in our scenario.

\section{Experiments}

We conduct various experiments to validate the effectiveness of Robusta and our methods. One such experiment, described in section \ref{ssec:exp1quality}, demonstrates that the images generated by Robusta are of comparable quality to those produced by the state-of-the-art (SOTA) label-to-image translation. For a more detailed discussion on this topic, please refer to Appendix~\textbf{C}. In addition, we assess the quality of Robusta-generated images under different input corruptions in section \ref{ssec:exp2RobustnessGAN}. We also evaluate the impact of using Robusta-generated data on the robustness of semantic segmentation algorithms and OOD detection in section \ref{ssec:exp3Robustnessaleatoric}.%

\subsection{Quality of Robusta's images}\label{ssec:exp1quality}

To evaluate the quality of the images generated by our cGAN-cascade, we adopt the evaluation protocol used in previous studies on label-to-image translation \cite{sushko2021oasis, spadepark2019semantic}. Specifically, we convert label maps into RGB synthetic images and measure the FID and mIoU, expressed in \%. We report experimental details and exhaustive comparisons with current SOTA methods and baselines \cite{spadepark2019semantic, le2020hierarchicalCRN, qi2018sims, liu2019learningCCFPSE, tang2020localLGGAN}, including OASIS \cite{sushko2021oasis}, on multiple datasets \cite{caesar2018cocostuff, cordts2016cityscapes, zhou2017scene, jeong2019kaist, spadepark2019semantic} in Appendix~\textbf{C}. Our experiments show that we achieve equivalent or superior performance compared to current 
methods on most datasets. We also provide qualitative assessments of the synthesized images in Appendix~\textbf{G} using numerous visual examples to provide a more comprehensive evaluation.

\subsection{Robustness of the cGAN cascade}\label{ssec:exp2RobustnessGAN}

We evaluate the robustness of our pipeline using the protocol proposed in Section \ref{ssec:eval_protocol_robust}  on Cityscapes~\cite{cordts2016cityscapes} and ADE20K~\cite{zhou2017scene}%
. The results are summarized in Table \ref{table:res_robustness_GAN} and show that our GAN-cascade performs at least as well as other SOTA methods on most datasets. Qualitatively, in Figure \ref{fig:qualitative_robustness}, we observe that the SOTA algorithms struggle to generate images when given perturbed label maps, indicating that these algorithms overfit strongly to their training dataset and lack robustness to changes in the input data. This highlights the importance of investigating the robustness of image-generation algorithms.

\vspace{-0.1cm}
\subsection{Evaluation of the segmentation robustness}\label{ssec:exp3Robustnessaleatoric}
\vspace{-0.1cm}
We conduct experiments over five datasets to assess the network's robustness to uncertainties. First, we do experiments on StreetHazards~\cite{hendrycks2019anomalyseg} and BDD anomaly~\cite{hendrycks2019anomalyseg} to evaluate the network's ability to detect OOD classes. The test set includes some object classes %
absent from the training set. To perform this task, we trained a deep neural network called Obsnet~\cite{besnier2021triggering} using the \emph{Outlier-StreetHazards} and \emph{Outlier-BDD anomaly} datasets generated by Robusta. We provide more information on these datasets in subsection \ref{subsubsec:OOD}. \emph{Furthermore}, we investigate the network's reliability for the aleatoric uncertainty by training a DNN on \emph{Corrupted-CS}, generated by Robusta. We then evaluate the network's performance on three datasets used for this task: Rainy~\cite{hu2019depth} and Foggy Cityscapes~\cite{sakaridis2018semantic}, and Cityscapes-C~\cite{hendrycks2019benchmarking}.

\noindent\textbf{Metrics.} We use several criteria to evaluate the performance of the segmentation models. The first criterion, the mIoU, measures the predictive performance of the %
networks in %
segmentation accuracy, as introduced by Jaccard~\cite{jaccard1912distribution}. Another criterion is the negative log-likelihood (NLL), which is a proper scoring rule based on the aleatoric uncertainty, similar to the approach described by Lakshminarayanan et al.~\cite{lakshminarayanan2017simple}. Additionally, we employ the expected calibration error (ECE)~\cite{guo2017calibration} to evaluate the grounding of the DNN's top-class confidence scores.
Furthermore, we assess the DNN's ability to detect OOD data using the AUPR, AUROC, and the False Positive Rate at 95\% recall (FPR95), as defined by Hendrycks et al.~\cite{hendrycks2016baseline}. By considering multiple metrics, we obtain a more comprehensive and precise assessment of the DNN's performance in %
accuracy, calibration error, failure rate, and OOD detection. Although it may be challenging to achieve top performance on all metrics, we argue that evaluating multiple metrics, as supported by research~\cite{ovadia2019can, fort2019deep}, is more practical and convincing than optimizing for a single metric at the potential expense of other factors.

\vspace{-0.2cm}
\subsubsection{Out-of-Distribution detection experiments}\label{subsubsec:OOD}
\vspace{-0.1cm}

\noindent\textbf{Datasets.} We conduct a study on outlier detection %
using StreetHazards~\cite{hendrycks2019anomalyseg}, BDD Anomaly~\cite{hendrycks2019anomalyseg} , and Road Anomaly~\cite{lis2019detecting, chan2021segmentmeifyoucan}. StreetHazards is a large-scale dataset %
comprising various sets of synthetic images of street scenes. It contains $5,125$ images for training and $1,500$ test images, with pixel-wise annotations for $13$ classes in the training set. The test set comprises $13$ training classes and $250$ out-of-distribution (OOD) classes unseen in the training set. This enables us to test the algorithm's robustness when facing diverse scenarios. The BDD Anomaly dataset is a subset of the BDD~\cite{yu2020bdd100k} dataset and contains 6,688 street scenes in the training set and 361 in the testing set. The training set includes 17 classes, and the test set consists of these 17 training classes as well as 2 OOD classes. Finally, RoadAnomaly contains 100 high-resolution test images of unusual dangers encountered in real life, such as giraffes, cattle, boats, etc. The images are associated with binary per-pixel labels for the background and the anomalies.

\noindent\textbf{Architecture.} In this experiment, we use Obsnet~\cite{besnier2021triggering} and DeepLabv3+~\cite{chen2018encoder} with a ResNet50~\cite{he2016deep} as backbone, following the protocol presented in~\cite{hendrycks2019anomalyseg}. Specifically, we adopt the evaluation criteria introduced by Hendrycks et al.~\cite{hendrycks2019anomalyseg}. Obsnet is a DNN designed to detect faults, and we train it using the \emph{Outlier-StreetHazards} and \emph{Outlier-BDD Anomaly} datasets generated by Robusta. We introduce additional OOD objects labeled as "cars" to the original label maps to create the new datasets and use Robusta to generate the corresponding images, as explained in Section \ref{sec:seg}. To evaluate Robusta's performance against other baselines, we also train a Binary segmentation model (denoted BCE + Robusta) to identify outliers using \emph{Outlier-BDD Anomaly} generated by Robusta. To ensure consistency with the other baselines, we employ the same DeepLabv3+ segmentation model.

\noindent\textbf{Baselines.} We evaluate our algorithm against several methods, including Deep Ensembles~\cite{lakshminarayanan2017simple}, BatchEnsemble~\cite{wen2020batchensemble}, LP-BNN~\cite{franchi2020encoding}, TRADI~\cite{franchi2019tradi}, MIMO~\cite{havasi2020training}, and Obsnet~\cite{besnier2021triggering}, on epistemic uncertainty. %
We also include the baseline - Maximum Class Probability (MCP) - using the maximum probability as a confidence score.

\noindent\textbf{Results.} We show in Table \ref{table:outofditribution} that our method outperforms all other methods on three out of four performance measures, achieving SOTA results on epistemic uncertainty. Moreover, our approach is faster than Deep Ensemble and LP-BNN, as it only requires a single inference pass, while the others need four. We also include results on RoadAnomaly~\cite{lis2019detecting, chan2021segmentmeifyoucan} in Table \ref{table:roadanom}, where we show that Robusta improves over standard Obsnet on the component-level metrics, all defined in the benchmark accompanying paper~\cite{chan2021segmentmeifyoucan}.

\begin{table}[t!]
\renewcommand{\figurename}{Table}
\begin{center}
\resizebox{\columnwidth}{!}{
\begin{tabular}{@{}lcccc<{\kern-\tabcolsep}}
\toprule
\multirow{2}{*}{\textbf{Method}} & \multicolumn{4}{c}{\textbf{StreetHazards}} \\
& \textbf{mIoU} $\uparrow$ & \textbf{AUROC} $\uparrow$  & \textbf{AUPR} $\uparrow$ & \textbf{FPR95} $\downarrow$ \\
\midrule
Baseline (MCP)~\cite{hendrycks2016baseline} & 53.90  & 86.60 & 6.91 & 35.74  \\ 
TRADI~\cite{franchi2019tradi} &  52.46	& 87.39 & 6.93  & 38.26\\
Deep Ensembles~\cite{lakshminarayanan2017simple}& \second 55.59 & 87.94 & 8.32 & 30.29 \\
MIMO~\cite{havasi2020training} & 55.44 &  87.38 &  6.90 & 32.66 \\ 
BatchEnsemble~\cite{wen2020batchensemble}  &\first\textbf{56.16}  & 88.17 & 7.59 & 32.85 \\ 
LP-BNN~\cite{franchi2020encoding}  &  54.50 & 88.33 & 7.18 & 32.61   \\
Obsnet~\cite{besnier2021triggering} & 53.90  & 94.96 & 10.58 & 16.74  \\
Obsnet + LA~\cite{besnier2021triggering} & 53.90  & 95.37 & 10.91 & 15.78  \\
BCE + Robusta (Ours) & 53.90 & \second 95.91 & \second 13.58 & \first\textbf{13.05}  \\
Obsnet + Robusta (Ours) & 53.90  & \first\textbf{96.27} & \first\textbf{15.60} & \second14.81  \\
\midrule 
& \multicolumn{4}{c}{\textbf{BDD Anomaly}} \\
\midrule
Baseline (MCP)~\cite{hendrycks2016baseline} &  47.63  & 85.15  & 4.50  & 28.78  \\ 
TRADI~\cite{franchi2019tradi}   & 44.26	& 84.80	& 4.54	& 36.87	\\ 
Deep Ensembles~\cite{lakshminarayanan2017simple}& \first\textbf{51.07}  & 84.80  & \second 5.24  & 28.55   \\
MIMO~\cite{havasi2020training}& 47.20  &  84.38  & 4.32  & 35.24  \\
BatchEnsemble~\cite{wen2020batchensemble}  &  48.09  & 84.27  & 4.49  & 30.17    \\ 
LP-BNN~\cite{franchi2020encoding}    & \second 49.01  &85.32  & 4.52  & 29.47   \\ 
Obsnet~\cite{besnier2021triggering}  & 47.63 & 87.66 & 1.01 & \second 19.50   \\
Obsnet + LAA~\cite{besnier2021triggering} & 47.63 & 88.16 & 1.71 &21.71  \\
BCE + Robusta (Ours) & 47.63 & \second 92.99 & 2.45 & 20.06  \\
Obsnet + Robusta (Ours) & 47.63 & \first\textbf{96.86} & \first\textbf{5.53} & \first\textbf{14.51} \\
\bottomrule
 \end{tabular}
 } %
 \end{center}
 \caption{\textbf{Comparative results on the OOD task for semantic segmentation.} The architecture is a DeepLabv3+ based on ResNet50.}\label{table:outofditribution}
 \vspace{-2mm}
\end{table}
 
\begin{table}[t!]
\renewcommand{\figurename}{Table}
\begin{center}
\resizebox{\columnwidth}{!}{
\begin{tabular}{@{}lccccc<{\kern-\tabcolsep}}
\toprule
\textbf{Method} & \textbf{AUPR} $\uparrow$ & \textbf{FPR95} $\downarrow$  & \textbf{SIoUgt} $\uparrow$ &  \textbf{PPV} $\downarrow$ & \textbf{mF1} $\uparrow$\\
 \midrule 
Baseline (MCP)~\cite{hendrycks2016baseline} &  28.0  & 72.1 &  15.5 &  15.3 & 5.4 \\ 
Deep Ensembles~\cite{lakshminarayanan2017simple} &  17.7  & 91.1  & 16.4  & 20.8 & 3.4 \\
Obsnet + LAA~\cite{besnier2021triggering} & \first\textbf{75.4} & \first\textbf{26.7} & \second 44.2 &\second 52.6 &\second 45.1 \\
BCE + Robusta (Ours) & \second 70.3 & \second 45.3 &  \first\textbf{48.2} & \first\textbf{57.6} &  \first\textbf{48.2} \\
\bottomrule
 \end{tabular}
}
\end{center}
\caption{\textbf{Comparative results on RoadAnomaly~\cite{lis2019detecting}.} These methods are trained without any supplemental OOD data.}\label{table:roadanom}
\vspace{-2mm}
\end{table}
\vspace{-0.2cm}
\subsubsection{Aleatoric uncertainty experiments}\label{aleatoric}
\vspace{-0.2cm}
The objective of our study is to evaluate the ability of DNNs to handle aleatoric uncertainty in semantic segmentation. We utilize Rainy~\cite{hu2019depth} and Foggy Cityscapes~\cite{sakaridis2018semantic}, which involve the introduction of rain or fog to the Cityscapes validation dataset. We generate aleatoric uncertainties on the Cityscapes validation set to create Cityscapes-C~\cite{michaelis2019benchmarking} with varying levels of perturbations using different methods such as Gaussian noise, shot noise, impulse noise, defocus blur, frosted, glass blur, motion blur, zoom blur, snow, frost, fog, brightness, contrast, elastic, pixelate, and JPEG. We used the code of Hendrycks et al.~\cite{hendrycks2019benchmarking} to generate these perturbations.

To assess the reliability of the DNNs in the presence of aleatoric uncertainty, we measure the ECE, mIoU, and NLL in Table \ref{table:Aleatoric}. We obtain results comparable to %
current methods, with Deep Ensembles performing better than the other approaches. However, Deep Ensembles require training multiple DNNs, which is more time-consuming for both training and inference. 

\begin{table}[t!]
\begin{center}
\resizebox{\columnwidth}{!}{
\begin{tabular}{@{}lccccccc<{\kern-\tabcolsep}}
\toprule
\multirow{2}{*}{\textbf{Method}} & \multicolumn{3}{c}{\textbf{Cityscapes}} & \multicolumn{3}{c}{\textbf{Rainy Cityscapes}}   \\
    &  \textbf{mIoU} $\uparrow$ &  \textbf{ECE} $\downarrow$  &  \textbf{NLL} $\downarrow$  &  \textbf{mIoU} $\uparrow$  &  \textbf{ECE} $\downarrow$ &  \textbf{NLL} $\downarrow$    \\
\midrule
Baseline (MCP)~\cite{hendrycks2016baseline} &  76.51 & 13.03 &-0.9456 & 58.98 & 13.95  & -0.8123\\
MIMO  \cite{havasi2020training}  &  77.13 & 13.98  & \second-0.9516 & 59.27 & 14.36 & -0.8135\\
BatchEnsemble \cite{wen2020batchensemble} & \second 77.99 & \second 11.29 &-0.9472 & 60.29 & 14.36 &-0.7820\\
LP-BNN  \cite{franchi2020encoding}  &  77.39 & \first\textbf{11.05}  &-0.9464 & \second60.71 & 13.38 &-0.7891\\
Deep Ensembles  \cite{lakshminarayanan2017simple} &  77.48 & 12.74 &-0.9469 & 59.52  &\first\textbf{10.78}   &\second-0.8205 \\
MCP +   Robusta (Ours)  &   \first\textbf{78.41} &	12.11	 & \first\textbf{-0.9546}	 & \first\textbf{62.31} & \second 12.54  & \first \textbf{-0.8382} \\
\midrule
&\multicolumn{3}{c}{\textbf{Foggy Cityscapes}}  & \multicolumn{3}{c}{\textbf{Cityscapes-C}} \\
\midrule
Baseline (MCP)~\cite{hendrycks2016baseline} &  69.89  & 14.93 & -0.9001  & 40.85   & 22.42 & -0.7389 \\
MIMO  \cite{havasi2020training}  &  70.24 & 14.25 &-0.9014& 40.73    & 23.50 &-0.7313\\
BatchEnsemble \cite{wen2020batchensemble} & \second 72.19  & 14.25 &\first \textbf{-0.9132}&   40.93 &  22.70 &-0.7082\\
LP-BNN  \cite{franchi2020encoding}  &\first \textbf{72.39} &\second 13.58 &\first\textbf{-0.9131}& \second43.47    & \second20.85 &-0.7282\\
Deep Ensembles  \cite{lakshminarayanan2017simple} & 71.43 & \second 14.07  &-0.9070 & 43.40  & \first\textbf{19.12}&\second-0.7509 \\
MCP +   Robusta (Ours)  &  	72.01 &	14.38 & -0.9110	& \first\textbf{43.49} & 24.97 & \first\textbf{-0.7578} \\
\bottomrule
\end{tabular}
}
\end{center}
\caption{\textbf{Evaluation of the influence of the quality of the training dataset.} We trained Deeplabs V3+ with ResNet101 backbones %
on CS (and CS + Corrupted-CS for MCP + Robusta) %
and tested the models 
on different corrupted variants of the original validation set.%
}
\label{table:Aleatoric}
\vspace{-4mm}
\end{table}

\vspace{-0.1cm}
\section{Discussions and details}
\vspace{-0.1cm}
A comprehensive exposition of Robusta, including its implementation details, can be found in Appendices \textbf{A} and \textbf{B}. Appendix \textbf{C} evaluates Robusta as a label-to-image translation model and compares its performance with other state-of-the-art algorithms. Appendix \textbf{D} provides additional details on morphology and the \emph{morphological} dataset derivation process. Extensive ablation experiments are conducted in Appendix \textbf{E} to demonstrate the significance of each contribution of Robusta. Finally, Appendix \textbf{F} highlights the trade-off between computational cost and performance.

\vspace{-0.1cm}
\section{Conclusion}
\label{sec:Conclusion}
\vspace{-0.15cm}
In this work, we propose a novel approach to enhancing the robustness of deep semantic segmentation networks by generating new textures and outlier objects from existing label maps. For best performance, we propose a new robust label-to-image model called Robusta, which incorporates Transformers and two conditional GANs. We verify that it is able to produce high-quality images even when asked to generate outliers, thanks to a new robustness evaluation framework for generative models, on which Robusta performs better than current methods. 

Furthermore, we demonstrate that using the datasets generated with Robusta during training helps improve the robustness of semantic segmentation algorithms. We show significant improvements in the performance of segmentation models when given perturbed inputs as well as for out-of-distribution detection.

\vspace{-0.15cm}
\section*{Acknowledgments}
\vspace{-0.25cm}
This work was supported by AID Project ACoCaTherm and performed using HPC resources from GENCI-IDRIS (Grant 2022-AD011011970R2). AB was supported by ELSA project funded by the European Union under grant agreement No. 101070617.

\clearpage

{
\small
\bibliographystyle{ieee_fullname}
\bibliography{egbib}
}

\renewcommand{\theequation}{S\arabic{equation}}
\renewcommand{\thefigure}{S\arabic{figure}}
\renewcommand{\thetable}{S\arabic{table}}
\renewcommand{\thesection}{\Alph{section}}
\newcommand{\rotr}[1]{\rotatebox{90}{#1}}

\appendix
\startcontents
\twocolumn[{%
\renewcommand\twocolumn[1][]{#1}%
\renewcommand\contentsname{\LARGE \centering Table of Contents - Supplementary Material}
\vspace{1cm}
\addtocontents{toc}{\protect\setcounter{tocdepth}{2}}
{
\hypersetup{linkcolor=black}
\printcontents{ }{1}{\section*{\contentsname}}{}
\vspace{8mm}
}
}]

\clearpage

\section{From to pix2pixHD to Robusta}\label{sec:Baselines_appendix}

\subsection{Background on label-to-image translation methods}
Conditional Generative Adversarial Networks are one of the most popular tools for semantic image synthesis in the literature~\cite{pang2021image}. These models consist of generators with conditioning designed to control the models and generate specific contents. The standard loss function for cGAN training is defined as follows:

\begin{multline}\label{eq:GANloss_supp}
\mathcal{L}_{cGAN}(\vtheta_G, \vtheta_D) =\mathbb{E}_{\vx}[\log D(\vx \mid \vy)]+ \\ \mathbb{E}_{\vz}[\log (1-D(G(\vz \mid \vy)))],
\end{multline}

\noindent where $\vy$ is the conditional information and $\vtheta_G$ and $\vtheta_D$, represent the weights of the generator and the discriminator, respectively. In label-to-image translation, $\vy$ is the input label, $\vx$ is the target RGB image, and $\vz$ is a sampled latent variable.

\begin{figure*}[t]
\includegraphics[width=0.8\linewidth]{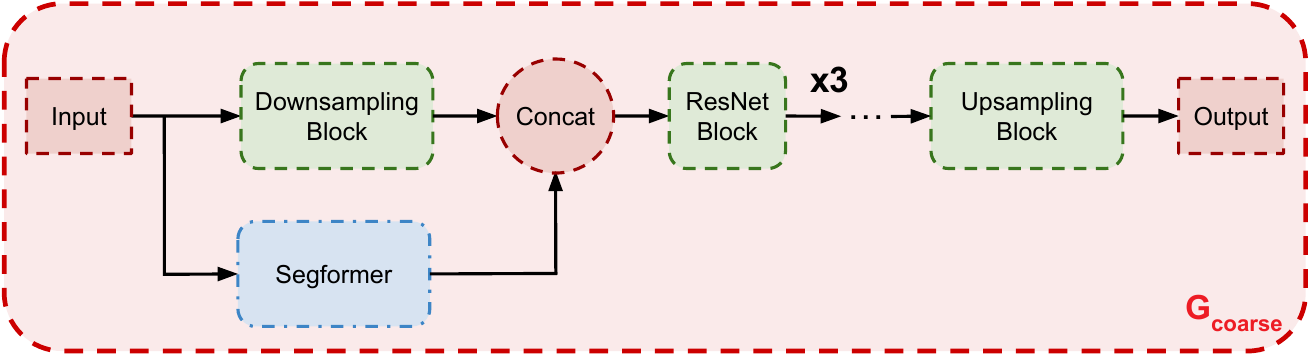}
\centering
\caption{\textbf{Illustration of the first generator of Robusta, $G_{\text{coarse}}$.} The green blocks are detailed in Table \ref{tab:robusta_details}.}
\label{fig:robusta_details}
\end{figure*}

\paragraph{Pix2pix~\cite{isola2017image}} is a type of cGAN specialized in image-to-image translation tasks. It uses a U-Net~\cite{ronneberger2015u} as the backbone and a PatchGAN~\cite{isola2017image} as the discriminator. In addition to the adversarial loss defined in Eq.\eqref{eq:GANloss_supp}, an L1 loss, defined in Eq.\eqref{eq:LL1}, is added to the cost function of cGANs to reduce blur:

\begin{equation}\label{eq:LL1}
\mathcal{L}_{L 1}(\vtheta_G)=\mathbb{E}_{(\vx, \vy, \vz)} [ \|\vx-G(\vz \mid \vy)\|_1].
\end{equation}

The total cost function of pix2pix is a linear combination of $\mathcal{L}_{L 1}$ and $\mathcal{L}_{cGAN}$, with $\lambda$ as a regularization hyper-parameter that balances the cGAN and reconstruction losses:

\begin{equation}
\label{eq:Lpix2pix}
\mathcal{L}_{\text{pix2pix}}(\vtheta_G, \vtheta_D)=\mathcal{L}_{cGAN}(\vtheta_G, \vtheta_D)+\lambda \mathcal{L}_{L 1}(\vtheta_G).
\end{equation}

\paragraph{Pix2pixHD~\cite{wang2018high}} improves the quality of the generated images thanks to enhanced multi-scale generators and discriminators. Pix2pixHD splits the generator and discriminator into two subGANs: $G_{1}$ and $D_1$, and $G_{2}$ and $D_2$. The loss function consists of the pix2pix loss function completed by a
\emph{feature-matching loss} and a \emph{perceptual loss}. The feature-matching loss $\mathcal{L}_{\text{FM}}(\vtheta_G, \vtheta{D_k})$ quantifies the distances between the feature maps of the real image $\vx$ and the predicted image. The feature maps are extracted from discriminator layers denoted by $i$. The loss is defined as follows:

\begin{multline}
\mathcal{L}_{\mathrm{FM}}\left(\vtheta_G, \vtheta_{D_k} \right) = \\
\mathbb{E}_{(\vx, \vy, \vz)} \sum_{i=1}^I \frac{1}{N_i}\left\|D_k^{(i)}(\vx)-D_k^{(i)}( G(\vz \mid \vy))\right\|_1,
\end{multline}
where $D_k^{(i)}$ stands for the $i$th-layer feature extractor of discriminator $D_k$, $I$ is the total number of layers, and $N_i$ represents the number of elements in each layer. The perceptual loss is designed to measure the similarity between the high-level features of the generated and real images. The feature maps are extracted from the $i$-th layer of a VGG network~\cite{simonyan2014very}, $F_{\text{VGG}}$, pretrained on ImageNet~\cite{deng2009imagenet}. The perceptual loss is defined as:

\begin{multline}
\mathcal{L}_{\mathrm{VGG}}(\vtheta_G)= \\
\mathbb{E}_{(\vx, \vy, \vz)}\sum_{i=1}^N \frac{1}{M_i}\left\|F_{\text{VGG}}^{(i)}(\vx)-F_{\text{VGG}}^{(i)}(G(\vz \mid \vy))\right\|_1,
\end{multline}
where $M_i$ is the number of elements of the VGG network, and $I$ is the number of layers on which we extract the feature maps.

\paragraph{SPADE~\cite{spadepark2019semantic}}
introduces a novel conditional normalization layer called SPatially-Adaptive (DE)normalization. In the context of image synthesis, Park et al. \cite{spadepark2019semantic} highlight that deeper layers of DNNs can easily lose semantic information due to the relative sparsity of semantic input. Since semantic inputs exhibit low local variance, common normalization methods like Batch Normalization \cite{ioffe2015batch} can inadvertently remove semantic information. Instance Normalization \cite{ulyaniov2016instance} can be used to avoid this issue. Notably, these normalization layers involve two steps: normalization and denormalization. In SPADE, Park et al.~\cite{spadepark2019semantic} propose learning denormalization factors at the pixel level, which are dependent on the label image $\vy$. Therefore, the output at site ($n$, $c$, $h$, $w$) of the $i$-th SPADE layer is defined as:

\begin{equation}
\gamma_{c, h, w}^{(i)}(\vy) \frac{a_{n, c, h, w}^{(i)} - \mu_c^{(i)}}{\sigma_c^{(i)}} + \beta_{c, h, w}^{(i)}(\vy),
\end{equation}
where $\mu_c^{(i)}$ and $\sigma_c^{(i)}$ represent the mean and variance at the instance or batch level. The activation $a$ is normalized by subtracting the mean and dividing by the standard deviation, followed by a modulation using the learned parameters $\gamma$ and $\beta$. Finally, $n$, $c$, $h$, and $w$ represent the number of the image in the batch, the channel, height, and width of the image, respectively.

\subsection{Robusta's loss functions}
\label{sec:Robusta_losses}

To train our model, we divide the training procedure into two parts.
We begin by training the first generator $G_{\text{coarse}}$ using the original label maps to produce synthesized images. Next, we use these synthesized images to feed and train the second generator $G_{\text{fine}}$.

\paragraph{ $G_{\text{coarse}}$ - loss.}
We use the same loss functions as in Wang et al.~\cite{wang2018high}, namely $\mathcal{L}_{\text{pix2pix}}$, $\mathcal{L}_{\text{FM}}$ and $\mathcal{L}_{\text{VGG}}$:

\begin{equation}
    \mathcal{L}_{\text{coarse}} = \mathcal{L}_{\text{pix2pix}} + \lambda_{\text{FM}}\mathcal{L}_{\text{FM}} + \lambda_{\text{VGG}}\mathcal{L}_{\text{VGG}} %
\end{equation}

\paragraph{ $G_{\text{fine}}$ - loss.}
After training $G_{\text{fine}}$, we generate low-resolution images $\textit{I}_{\text{LR}}$ to be used in the training of $G_{\text{fine}}$. We use the Least Squares Generative Adversarial Network approach, as suggested in SRGAN \cite{ledig2017photo}, which yields the following loss function:

\begin{dmath}\label{eq:LSGANloss}
\mathcal{L}_{\text{cGAN}}^{\text{MSE}}(\vtheta_G, \vtheta_D) =\mathbb{E}_{\textit{I}_{\textit{HR}} \sim p_{\textit{I}_{\textit{HR}}}}[\left(1 - D_{\vtheta_D}(\textit{I}_{\textit{HR}})\right)^2] + \mathbb{E}_{\textit{I}_{\textit{LR}} \sim p_{\textit{I}_{\textit{LR}}}}[D_{\vtheta_D}(G_{\vtheta_G}(\textit{I}_{\textit{LR}}))^2].
\end{dmath}

\noindent
The final loss
for training $G_{\text{fine}}$
reads:

\begin{equation}
     \mathcal{L}_{\text{fine}}  = \mathcal{L}_{\text{cGAN}}^{\text{MSE}} +
    \lambda_{\textit{FM}}\mathcal{L}_{\textit{FM}} + \lambda_{\textit{VGG}}\mathcal{L}_{\textit{VGG}} %
\end{equation}

\noindent
The hyper-parameters used to train both $G_{\text{coarse}}$ and $G_{\text{fine}}$ are reported in Appendix \ref{ssec:hypers}.

\subsection{Details on Robusta's architecture}

Figure \ref{fig:robusta_details} and Table \ref{tab:robusta_details} detail the architecture and technical specifications on the first generator of Robusta, $G_{\text{coarse}}$.

\begin{table}[t]
\centering
\scalebox{0.84}
{

\begin{tabular}{|c|}
\hline
\rowcolor[HTML]{D9EAD3}
\multicolumn{1}{|c|}{\cellcolor[HTML]{D9EAD3}\textbf{Downsampling Block}} \\ \hline
Conv2D(in\_ch=35, out\_ch=64, kernel\_size=7)                             \\
Batch\_Norm(64)                                                           \\
ReLU                                                                      \\
Conv2D(in\_ch=64, out\_ch=128, kernel\_size=3, stride=2)                  \\
Batch\_Norm(128)                                                          \\
ReLU                                                                      \\ \hline
\rowcolor[HTML]{D9EAD3}
\multicolumn{1}{|c|}{\cellcolor[HTML]{D9EAD3}\textbf{ResNet Block}}       \\ \hline
Conv2D(in\_ch=256, out\_ch=256, kernel\_size=3)                           \\
SPADE(256)                                                                \\
ReLU                                                                      \\
Conv2D(in\_ch=256, out\_ch=256, kernel\_size=3)                           \\
SPADE(256)                                                                \\ \hline
\rowcolor[HTML]{D9EAD3}
\multicolumn{1}{|c|}{\cellcolor[HTML]{D9EAD3}\textbf{Upsampling Block}}   \\ \hline
ConvTranspose2D(in\_ch=256, out\_ch=128, kernel\_size=3, stride=2)        \\
SPADE(128)                                                                \\
ReLU                                                                      \\
Conv2D(in\_ch=128, out\_ch=3, kernel\_size=3)                             \\
TanH                                                                      \\  \hline
\end{tabular}
}
\caption{
\textbf{Technical details on $G_{\text{coarse}}$.}
}
\label{tab:robusta_details}
\end{table}

\paragraph{Why splitting $G_{\text{coarse}}$ into two modules?}
Robusta involves changing one of the submodules from $G_{\text{coarse}}$ with a Segformer~\cite{xie2021segformer}. While we keep the global structure of pix2pixHD~\cite{wang2018high} with a GAN generating outputs to be concatenated in the latent space of the main generative network, the philosophy is completely different. The objective of pix2pixHD was to be able to work at different resolutions, possibly with a greater number of submodules. In our case, the objective of the Segformer submodule is to leverage two different architectures, which are known to focus on different aspects of the data~\cite{park2022how},
and fuse their learned information by concatenation.
For instance, transformers~\cite{dosovitskiy2020image} tend to learn more about the context and low frequencies in the images. With this in mind, we use the same input resolution for the transformer and its convolutional counterpart, as opposed to pix2pixHD.

\paragraph{The interface between the two modules}

We use a dense layer to project the output of the Segformer submodule to the size of the latent space of its convolutional counterpart to enable the concatenation of the activations. We also adapt the number of channels to those of the latent activation of the convolutional network using the
$\texttt{fuse-conv}$ proposed by Xie et al.\cite{xie2021segformer}.

\paragraph{The usefulness of SPADE layers}

We incorporate SPADE in all the batch normalization layers of the ResNet and Upsampling Block of $G_{\text{coarse}}$ to reinject spatial information. Specifically, we use SPADE after the convolutions that come after the concatenation of the Segformer output and the latent space of the convolutional network.

\clearpage

\section{Implementation details}\label{app:details}

\subsection{Training hyperparameters}\label{ssec:hypers}

This section includes the hyper-parameters utilized in the label-to-image translation, semantic segmentation, and anomaly detection experiments. These hyper-parameters are presented in Tables \ref{table:suptab1}, \ref{table:suptab2}, and \ref{table:suptab3}. Our implementation is based on PyTorch~\cite{paszke2019pytorch}.

\begin{table*}[t]
\begin{center}
\scalebox{1}
{
\begin{tabular}{lccccc}
\toprule
  & \multicolumn{1}{c}{ADE20K}  & \multicolumn{1}{c}{AO}      & \multicolumn{1}{c}{Cityscapes} & \multicolumn{1}{c}{COCO-stuff} & KAIST   \\ \hline
Segformer                & \multicolumn{1}{c}{B5}               & \multicolumn{1}{c}{B5}               & \multicolumn{1}{c}{B5}                  & \multicolumn{1}{c}{B0}                  & B5               \\ \hline
Learning rate            & \multicolumn{1}{c}{0.0002}           & \multicolumn{1}{c}{0.0002}           & \multicolumn{1}{c}{0.0002}              & \multicolumn{1}{c}{0.0002}              & 0.0001           \\ \hline
Batch size               & \multicolumn{1}{c}{32}               & \multicolumn{1}{c}{16}               & \multicolumn{1}{c}{8}                   & \multicolumn{1}{c}{8}                   & 32               \\ \hline
Resolution        & \multicolumn{1}{c}{$286 \times 286$} & \multicolumn{1}{c}{$256 \times 256$} & \multicolumn{1}{c}{$512 \times 256$}    & \multicolumn{1}{c}{$286 \times 286$}    & $256 \times 256$ \\ \hline
Weight decay             & \multicolumn{1}{c}{0}                & \multicolumn{1}{c}{0}                & \multicolumn{1}{c}{0.00001}             & \multicolumn{1}{c}{0}                   & 0                \\ \hline
Epochs                   & \multicolumn{5}{c}{200}                                                                                                                                                                \\ \hline
Random crop              & \multicolumn{1}{c}{$256 \times 256$} & \multicolumn{1}{c}{$\varnothing$}    & \multicolumn{1}{c}{$\varnothing$}       & \multicolumn{1}{c}{$256 \times 256$}    & $\varnothing$    \\ \hline
$\lambda_{\textit{VGG}}$ & \multicolumn{5}{c}{5}                                                                                                                                                                  \\ \hline
$\lambda_{\textit{FM}}$  & \multicolumn{5}{c}{5}                                                                                                                                                                  \\ \bottomrule
\end{tabular}
}
\end{center}
\caption{
\textbf{Hyperparameter configuration used in the training of $G_{\text{coarse}}$.} AO is ADE20K-outdoors.
}\label{table:suptab1}
\end{table*}

\begin{table*}[t]
\begin{center}
\scalebox{1}
{
\begin{tabular}{lccccc}
\toprule
                           & \multicolumn{1}{c}{ADE20K}             & \multicolumn{1}{c}{AO}                 & \multicolumn{1}{c}{Cityscapes}         & \multicolumn{1}{c}{COCO-stuff}         & KAIST              \\ \hline
Learning rate              & \multicolumn{5}{c}{0.0002}                                                                                                                                                                 \\ \hline
Batch size                 & \multicolumn{1}{c}{12}                 & \multicolumn{1}{c}{12}                 & \multicolumn{1}{c}{8}                  & \multicolumn{1}{c}{10}                 & 10                 \\ \hline
Resize resolution          & \multicolumn{1}{c}{\(256 \times 256\)} & \multicolumn{1}{c}{\(256 \times 256\)} & \multicolumn{1}{c}{\(512 \times 256\)} & \multicolumn{1}{c}{\(286 \times 286\)} & \(256 \times 256\) \\ \hline
Weight decay               & \multicolumn{5}{c}{0}                                                                                                                                                                      \\ \hline
Epochs                     & \multicolumn{5}{c}{100}                                                                                                                                                                    \\ \hline
Random crop                & \multicolumn{1}{c}{$\varnothing$}      & \multicolumn{1}{c}{$\varnothing$}      & \multicolumn{1}{c}{$\varnothing$}      & \multicolumn{1}{c}{\(256 \times 256\)} & $\varnothing$      \\ \hline
\(\lambda_{\textit{VGG}}\) & \multicolumn{1}{c}{1}                  & \multicolumn{1}{c}{10}                 & \multicolumn{1}{c}{10}                 & \multicolumn{1}{c}{1}                  & 10                 \\ \hline
\(\lambda_{\textit{FM}}\)  & \multicolumn{1}{c}{0.1}                & \multicolumn{1}{c}{1}                  & \multicolumn{1}{c}{1}                  & \multicolumn{1}{c}{1}                  & 1                  \\ \bottomrule
\end{tabular}
}
\end{center}
\caption{
\textbf{Hyperparameter configuration used in the training of $G_{\text{fine}}$.} AO is ADE20K-outdoors.
  }\label{table:suptab2}
\end{table*}

\begin{table*}[t]
\begin{center}
\scalebox{1}
{
\begin{tabular}{lccccc}
\toprule
                                                & \multicolumn{3}{c}{\textbf{Perturbation robustness}}                                        & \multicolumn{2}{c}{\textbf{Outlier detection}}               \\
 & \multicolumn{1}{c}{StreetHazards} & \multicolumn{1}{c}{BDD Anomaly} & Cityscapes  & \multicolumn{1}{c}{StreetHazards} & BDD Anomaly \\ \hline
Architecture                                    & \multicolumn{3}{c}{Deeplab v3+} & \multicolumn{2}{c}{Deeplab v3+}   \\ \hline
Backbone                                        & \multicolumn{2}{c}{ResNet 50}  & ResNet 101 & \multicolumn{2}{c}{ResNet 50}             \\ \hline
Output stride                                   & \multicolumn{3}{c}{8}  & \multicolumn{2}{c}{8}             \\ \hline
Learning rate                                   & \multicolumn{2}{c}{0.02}        &   0.1      & \multicolumn{2}{c}{0.002}   \\ \hline
Batch size                                      & \multicolumn{2}{c}{4}         & 8           & \multicolumn{2}{c}{6}  \\ \hline
Epochs                                          & \multicolumn{2}{c}{25}     & 80          & \multicolumn{2}{c}{15}    \\ \hline
Weight decay                                    & \multicolumn{3}{c}{0.0001}     & \multicolumn{2}{c}{0}   \\ \hline
Random Crop                                     & \multicolumn{2}{c}{$\varnothing$}            & \multicolumn{1}{c}{(768, 768)}        & \multicolumn{2}{c}{(80, 150)}      \\ \bottomrule
\end{tabular}
}
\end{center}
\caption{
 \textbf{Hyper-parameter configuration used in semantic segmentation experiments.}
}
\label{table:suptab3}
\end{table*}

\section{Quality of the images generated by Robusta}

\subsection{Datasets, metrics \& baselines}
\subsubsection{Datasets.}
We performed experiments on well-established datasets in label-to-image translation: ADE20K~\cite{zhou2017scene}, COCO-stuff~\cite{caesar2018cocostuff}, Cityscapes~\cite{cordts2016cityscapes}, KAIST~\cite{jeong2019kaist}, and ADE20K-outdoors~\cite{spadepark2019semantic}, a subset of ADE20k containing only outdoor scenes. In all our label-to-image translation experiments, the images are resized at the resolution of $256\times 256$, except for Cityscapes, for which the resolution is set to $256\times 512$.

\subsubsection{Performance metrics.}\label{performance_metrics}
Following common protocol from prior label-to-image translation works \cite{sushko2021oasis,spadepark2019semantic}, we assess
the quality of the generated images with two primary metrics: the Fréchet Inception Distance (FID) and the mean intersection over union (mIoU). FID measures the similarity of two sets of images based on their visual features, extracted by a pre-trained Inceptionv3 model \cite{szegedy2015going}. Lower scores correspond to better visual similarity between the sets of images. The mean Intersection over Union is used to evaluate the accuracy of semantic segmentation DNN's prediction and, therefore, provides hints on the quality of the rendering. In addition, we evaluate the quality of OOD detection on Outlier-Cityscapes using the Areas Under the Precision/Recall curve (AUPR) and the operating Curve (AUROC), following Hendrycks and Gimpel \cite{hendrycks2019scaling}.

To evaluate mIoU, we require a pre-trained DNN capable of performing semantic segmentation. As in \cite{sushko2021oasis,spadepark2019semantic}, we use multi-scale DRN-D-105 \cite{yu2017dilated} for Cityscapes, DeepLabV2 \cite{chen2017deeplab} for COCO-Stuff, and UperNet101 \cite{xiao2018unified} for both ADE20K and ADE20K-outdoors datasets. For the KAIST dataset, which only contains bounding box annotations of persons, we evaluate the mean Average Precision (mAP) specifically for the person class, using a faster R-CNN model \cite{ren2015faster} with ResNet-101 backbone trained on the COCO dataset \cite{lin2014microsoft}.

\begin{table*}[t]
\setlength{\abovecaptionskip}{0.cm}

\begin{center}
 \scalebox{0.95} {
\begin{tabular}{@{}c|cc|cc|cc|cc|cc@{}}
 \toprule
		\multirow{2}{*}{\normalsize{} Method }  & \multicolumn{2}{c|}{\normalsize{} ADE20K~\cite{zhou2017scene}} & \multicolumn{2}{c|}{\normalsize{} AO~\cite{spadepark2019semantic}} & \multicolumn{2}{c|}{\normalsize{} Cityscapes~\cite{cordts2016cityscapes}} & \multicolumn{2}{c|}{\normalsize{} COCO-stuff~\cite{caesar2018cocostuff}}  & \multicolumn{2}{c}{\normalsize{} KAIST~\cite{jeong2019kaist}}
		\tabularnewline
		& \normalsize{} FID$\downarrow$  & \normalsize{} mIoU$\uparrow$ & \normalsize{} FID$\downarrow$  & \normalsize{} mIoU$\uparrow$ & \normalsize{} FID$\downarrow$  & \normalsize{} mIoU$\uparrow$ &  \normalsize{} FID$\downarrow$  & \normalsize{} mIoU$\uparrow$  &  \normalsize{} FID$\downarrow$  & \normalsize{} mAP$\uparrow$ \tabularnewline

		\midrule

		{\normalsize{} CRN }
		& \normalsize{73.3} & \normalsize{22.4}  & \normalsize{99.0} & \normalsize{16.5} &  \normalsize{104.7} & \normalsize{52.4} &  \normalsize{70.4} & \normalsize{23.7} &  \normalsize{n/a} & \normalsize{n/a} \tabularnewline

		{\normalsize{} SIMS  }
		& \normalsize{n/a} & \normalsize{n/a} & \normalsize{67.7} & \normalsize{13.1} &  \normalsize{49.7} & \normalsize{47.2}  &  \normalsize{n/a} & \normalsize{n/a}  &  \normalsize{n/a} & \normalsize{n/a} \tabularnewline

		{\normalsize{} pix2pixHD  }
		& \normalsize{81.8} & \normalsize{20.3} & \normalsize{97.8} & \normalsize{17.4} &  \normalsize{95.0} & \normalsize{58.3} &  \normalsize{111.5} & \normalsize{14.6}  &  \normalsize{55.4} & \normalsize{1.4} \tabularnewline

	 {\normalsize{} LGGAN  }
	 & \normalsize{31.6} & \normalsize{41.6} & \normalsize{n/a} & \normalsize{n/a} & \normalsize{57.7} & \normalsize{68.4} &  \normalsize{n/a} & \normalsize{n/a} &  \normalsize{n/a} & \normalsize{n/a} \tabularnewline

		{\normalsize{} CC-FPSE }
		& \normalsize{31.7} & \normalsize{43.7} &  \normalsize{n/a} & \normalsize{n/a} & \normalsize{54.3} & \normalsize{65.5} &  \normalsize{19.2} & \normalsize{41.6} &  \normalsize{n/a} & \normalsize{n/a} \tabularnewline

		\multirow{1}{*}{\normalsize{} SPADE }
		& \normalsize{33.9} & \normalsize{38.5} &  \normalsize{63.3} & \normalsize{30.8} &  \normalsize{71.8} & \normalsize{62.3} &  \normalsize{22.6} & \normalsize{37.4}  &  \normalsize{n/a} & \normalsize{n/a} \tabularnewline

	OASIS
	& \normalsize{28.3} & \normalsize{48.8} & \normalsize{48.6} & \normalsize{40.4} &  \normalsize{47.7} & \normalsize{69.3} & \normalsize{\textbf{17.0}} & \normalsize{\textbf{44.1}} &  \normalsize{n/a} & \normalsize{n/a} \tabularnewline

\midrule
	Robusta ($G_{\text{coarse}}$)
	& \normalsize{29.4} & \normalsize{46.7} & \normalsize{49.7} & \normalsize{40.6} &  \normalsize{56.3} & \normalsize{68.8} & \normalsize{30.4} & \normalsize{42.1} &  \normalsize{53.7} & \normalsize{\textbf{4.5}} \\

	Robusta $(G_{\text{coarse}}, G_{\text{fine}})$
	& \normalsize{\textbf{28.1}} & \normalsize{\textbf{49.0}} & \normalsize{\textbf{48.4}} & \normalsize{\textbf{41.8}} &  \normalsize{\textbf{47.1}} & \normalsize{\textbf{70.8}} & \normalsize{30.2} & \normalsize{42.8} &  \normalsize{\textbf{52.1}} & \normalsize{\textbf{4.5}}\\
 \bottomrule

		\end{tabular}

}
\end{center}
\vspace{-2mm}
\caption{\textbf{Comparison across datasets.} Our method outperforms the current leading methods in semantic segmentation (mIoU) and FID scores on most benchmark datasets. AO is ADE20K-outdoors.}
\label{table:RESULT_IQA}
\vspace{-2mm}
\end{table*}

\subsubsection{Baselines.}
In this study, we compare Robusta with Pix2PixHD \cite{wang2018high} and six other state-of-the-art baselines: SPADE~\cite{spadepark2019semantic}, CRN~\cite{le2020hierarchicalCRN}, SIMS~\cite{qi2018sims}, CC-FPSE~\cite{liu2019learningCCFPSE}, LGGAN~\cite{tang2020localLGGAN}, and OASIS~\cite{sushko2021oasis}. We either reproduce the results if the checkpoint is provided for each approach or retrain the models. All models are trained at the same resolutions to obtain the most accurate comparisons.

\subsection{Results of the experiments}
\subsubsection{Image quality results}
\label{app:sssec:GAN_quality}

We adopt the same evaluation protocol used in previous studies on label-to-image translation \cite{sushko2021oasis,spadepark2019semantic} to assess the quality of our images. Specifically, we convert label maps into RGB synthetic images and measure the FID and mIoU. Table \ref{table:RESULT_IQA} shows that we achieve equivalent or superior performance compared to state-of-the-art methods on most of the datasets.
Toward a more comprehensive evaluation, we also include qualitative assessments of the synthesized images in Appendix \ref{sec:qualitative} with multiple visual examples for each dataset.

\subsubsection{Domain adaptation results}

The last column of Table \ref{table:RESULT_IQA} presents the results for domain adaptation. To evaluate the effectiveness of our approach, we use an object detection DNN trained on RGB images and test its performance on infrared images using both Pix2PixHD \cite{wang2018high} and Robusta. Our results show that the mean average precision (mAP) on Pix2PixHD-generated images is only 1.40\%, but it significantly improves to 4.57\% with Robusta. These findings demonstrate our GAN cascade's efficacy, which improves label-to-image translation and enhances object detection performance across domains. For more details on the experimental protocol and instructions on using Robusta for this task, please refer to Table \ref{table:suptab2}.

\section{Background on morphology \& editing}

The field of mathematical morphology~\cite{serra1982image} is a well-established nonlinear image processing field that applies complete lattice theory to spatial structures.

Let $\vx$ be a grey-scale image with the intensity at position $p$ denoted as $\vx(p)$. Morphology involves two fundamental operations, grey-level dilation and grey-level erosion, which are defined as follows:

\begin{eqnarray}\label{ErosionDilation}
\varepsilon_{b}(\vx )(p)& = & \min_{h \in B} (\vx(p -h)- b(h)), \\
\delta_{b}(\vx)(p)     & = & \max_{h \in B} (\vx(p -h)+ b(h)).
\end{eqnarray}

In these equations, $\varepsilon_{b}(\vx)$ and $ \delta_{b}(\vx)$ represent morphological erosion and dilation, respectively, using a given structuring element $b$  of support $B$ \cite{serra1982image}. The structuring element's geometry determines the operators' effect, and for simplicity, we consider uniform structuring functions formalized by their support shape. More complex operators like opening and closing can be obtained by repeatedly applying these two basic operators.

Extending morphological operators to multivariate images, particularly color images or hyperspectral images, requires appropriate vector-ordering strategies. Researchers have developed various orders for multivariate images \cite{franchi2015ordering,velasco2011supervised,angulo2007morphological,cualiman2014probabilistic}. In this discussion, we focus on simplex ordering, like \cite{franchi2015ordering}, but propose a lexicographic ordering based on two components. First, we consider all non-instance pixels to be in the foreground, representing the smallest element of the order. Second, it is essential to order classes based on the editing task at hand. We recommend that classes representing small objects be assigned the highest values. Thus, for example, on Cityscapes, we would have the following ordering:

\small
$$
\mbox{traffic sign} >\mbox{traffic light} > \mbox{person} > \mbox{car} >\mbox{cycle} >\mbox{truck} >\mbox{train}.
$$
\normalsize

\begin{figure*}[t]
     \centering
     \begin{subfigure}[]{0.17\textwidth}
         \centering
         \includegraphics[width=\linewidth]{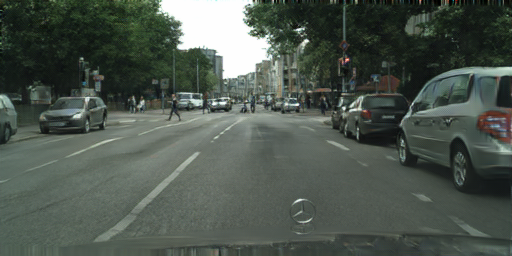}
         \caption{$\vx$}
     \end{subfigure}
     \begin{subfigure}[]{0.17\textwidth}
         \centering
         \includegraphics[width=\linewidth]{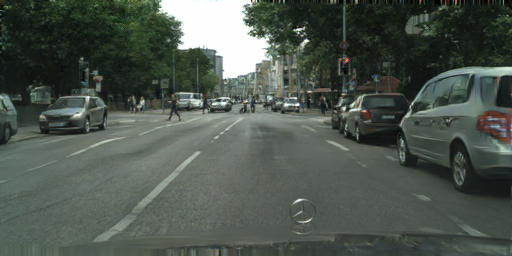}
         \caption{$\delta_{b_1}(\vx)$}
     \end{subfigure}
     \begin{subfigure}[]{0.17\textwidth}
         \centering
         \includegraphics[width=\linewidth]{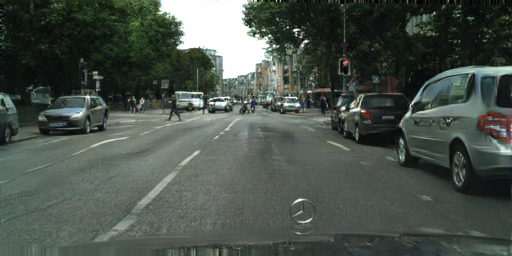}
         \caption{$\delta_{b_2}(\vx)$}
     \end{subfigure}
         \begin{subfigure}[]{0.17\textwidth}
         \centering
         \includegraphics[width=\linewidth]{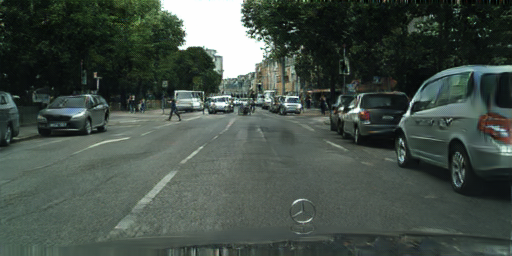}
         \caption{$\delta_{b_3}(\vx)$}
     \end{subfigure}
         \begin{subfigure}[]{0.17\textwidth}
         \centering
         \includegraphics[width=\linewidth]{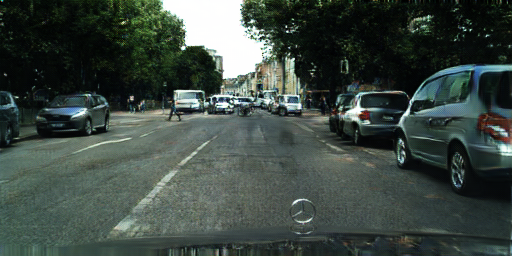}
         \caption{$\delta_{b_4}(\vx)$}
     \end{subfigure}
        \caption{\textbf{Illustration of the results of image morphological editing.} Figure (a) represents the results of label-to-image translation, and Figures (b)-(e) represent the  results of label-to-image translation applied to the dilated label maps with increasing structuring elements.}
        \label{fig:morph_editing_res}
\end{figure*}

\clearpage
\section{Ablation studies}

\subsection{Architectural choices}

\subsubsection{Ablation study on the architecture}

Table \ref{tab:architectural_contributions} shows the results of an ablation study for the different variations and improvements proposed by Robusta on two datasets - ADE20K-outdoors~\cite{spadepark2019semantic} (AO) and Cityscapes~\cite{cordts2016cityscapes}. Specifically, we measure the
impact of replacing one of the submodules of $G_{\text{coarse}}$
with a Segformer~\cite{xie2021segformer}, as well as the importance of using SPADE~\cite{spadepark2019semantic} layers in $G_{\text{coarse}}$, and $G_{\text{fine}}$, and the interaction with the new generator $G_{\text{fine}}$. The table reports FID and mIoU metrics, which indicate the quality of the generated images and the segmentation's accuracy, respectively. The table highlights the best-performing method corresponding to Robusta with all the proposed architectural contributions. It achieves the lowest FID and highest mIoU scores for both datasets. Specifically, it improves the FID by 19.4\% and the mIoU by 19.6\% on AO and 30.3\% and 13.4\% on CS for the FID and the mIoU, respectively.

Comparing the results, we can first observe that replacing the backbone of the submodule of $G_{\text{coarse}}$ significantly improves the performance of the model by 17\% of FID and 13.5\% of mIoU on AO and 11.1\% FID and 11.4\% mIoU on CS. Second, adding  $G_{\text{fine}}$ overall improves results and yields improvements, for instance, 3.2\% of mIoU on AO and 6\% of mIoU on CS, despite an increase of 3.5\% of the FID on AO and a decrease of 1\% of the mIoU on CS. Using SPADE layers also improves results overall.

In conclusion, the ablation study results show that adding architectural contributions such as the Segformer submodule, SPADE layers, and $G_{\text{fine}}$ can significantly improve the performance of the generating model for semantic segmentation tasks. Furthermore, the best results are obtained by using all the architectural contributions together in the Robusta model, which achieves the lowest FID and highest mIoU scores on both datasets. These results highlight the effectiveness of the proposed Robusta model for label-to-image tasks and demonstrate the importance of our architectural contributions to improve model performance.

\begin{table}[t]
\centering
\scalebox{0.7}
{
\begin{tabular}{@{}cccccccc<{\kern-\tabcolsep}}
\toprule
\multirow{2}{*}{SegFormer} & \multirow{2}{*}{SPADE 1} & \multirow{2}{*}{SPADE 2} & \multirow{2}{*}{$G_{\text{fine}}$} & \multicolumn{2}{c}{ADE20K-outdoors} & \multicolumn{2}{c}{Cityscapes} \\
& & & & FID $\downarrow$& mIoU $\uparrow$& FID $\downarrow$ & mIoU $\uparrow$\\
 \midrule

\xmark& \xmark& \xmark& \xmark &67.8&22.2&77.4&57.4\\
\cmark& \xmark& \xmark& \xmark & 50.8 & 35.7 & 56.3 & 68.8\\
\xmark& \cmark& \xmark& \xmark & 55.1 & 33.6 & 63.9 & 64.4\\
\xmark& \xmark& \xmark& \cmark & 52.3 & 30.5 & 67.5 & 60.9\\
\xmark& \cmark& \cmark& \cmark&52.4&37.7&53.6&67.5\\
\cmark& \xmark& \xmark& \cmark&54.3&38.9&50.0&67.8\\
\cmark& \cmark& \xmark& \cmark&49.6&39.2&47.9&66.1\\
\cmark& \xmark& \cmark& \cmark&56.5&41.0&48.0&68.3\\
\cellcolor{orange!20}\cmark& \cellcolor{orange!20}\cmark& \cellcolor{orange!20}\cmark& \cellcolor{orange!20}\cmark &\cellcolor{orange!20}\textbf{48.4}&\cellcolor{orange!20}\textbf{41.8}&\cellcolor{orange!20}\textbf{47.1}&\cellcolor{orange!20}\textbf{70.8}\\
 \bottomrule
\end{tabular}
}
\caption{\textbf{Ablation study of each architectural contribution}. The last row corresponds to Robusta. SPADE 1 is for SPADE in $G_{\text{coarse}}$. SPADE 2 is for SPADE in $G_{\text{fine}}$.}
\label{tab:architectural_contributions}
\end{table}

\subsubsection{Size of the Segformer}\label{section:AbblationSegFormer}

In this section, we investigate the influence of the size of the $G_{\text{coarse}}$ network on the performance of ($G_{\text{coarse}}$, $G_{\text{fine}}$) on Cityscapes and ADE20K. To do so, we experiment with different sizes of $G_{\text{coarse}}$ based on Segformer~\cite{xie2021segformer}, which offers five variants ranging from the lightest B0 to the heaviest B5. We report the results for all Segformer models in Table \ref{segformer_g_pix2pixhd}. Our findings show that increasing the complexity of the Segformer model leads to a higher mIoU score; however, this may not hold true for the FID. Additionally, we observe that the lightweight B0 variant performs well enough with lower complexity, which can be a good option for limited resources.

\begin{table}[t]
\centering
\scalebox{0.75}
{
\begin{tabular}{@{}c|cc|cc|cc@{}}
 \toprule
\multirow{2}{*}{Segformer} & \multicolumn{2}{c|}{ADE20K} & \multicolumn{2}{c|}{AO} & \multicolumn{2}{c}{Cityscapes} \\
& FID $\downarrow$& mIoU $\uparrow$& FID $\downarrow$ & mIoU $\uparrow$& FID $\downarrow$ & mIoU $\uparrow$\\
\midrule
B0   & 34.4 & 37.62 & 53.05 & 33.79 & 49.60 & 66.53 \\
B1   & 32.29 & 38.94 & 52.73 & 35.56 & 49.5 & 65.74  \\
B2 & 32.04 & 39.51 & 50.09 & 37.15 & 50.01 & 67.84 \\
B3   & 30.98 & 40.84 & 51.57 & 36.52 & 49.58 & 67.92 \\
B4 &  31.71 & 41.37 &50.80 &  36.89 & \textbf{48.56} & 68.14            \\
B5 &  \textbf{30.30} &  \textbf{43.93} & \textbf{49,65} & \textbf{40,58 } & 56.34 & \textbf{68.76} \\ %
 \bottomrule
\end{tabular}
}
\caption{\textbf{Performance of Robusta with different SegFormers~\cite{xie2021segformer}}. The largest Segformer, B5, yields the best results. AO is ADE20K-outdoors.}
\label{segformer_g_pix2pixhd}
\end{table}

\subsubsection{Usefulness of $G_{\text{fine}}$}\label{section:G3}

\begin{figure}[t]
\includegraphics[width=0.9\linewidth]{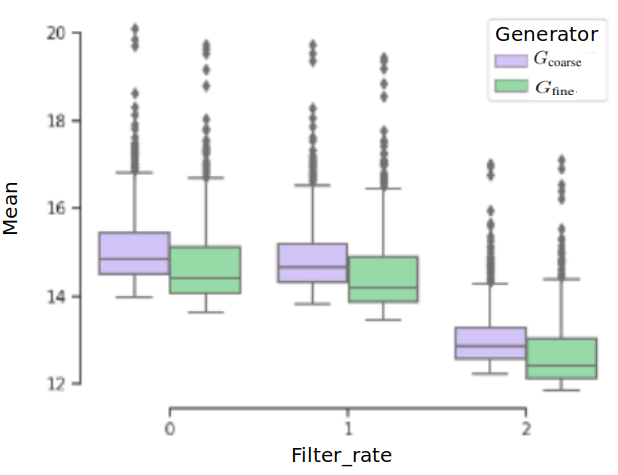}
\caption{\textbf{Frequency analysis of the usefulness of $G_{\text{fine}}$}. The filter rate represents the cutoff frequency of the high-pass filter. $\texttt{Filter\_rate} = 0 $, No filter. $\texttt{Filter\_rate} = 1$, Low cutoff frequency. $\texttt{Filter\_rate} = 2 $, High cutoff frequency.}
\label{fig:freqanalysis}
\vspace{-5pt}
\end{figure}

To evaluate the effectiveness of $G_{\text{fine}}$, we perform an analysis in the Fourier domain on 500 images from the Cityscapes validation set~\cite{cordts2016cityscapes}. For each image, we compute the fast Fourier transform of the ground truth, the output of $G_{\text{coarse}}$, and of $G_{\text{fine}}$. We then analyze the distance between the spectra of the generated images and the ground truth. Figure \ref{fig:freqanalysis} presents the mean distance among all the Fourier domains for three scenarios: one considering all the complete spectra and two considering only high frequencies.

As shown in Figure \ref{fig:freqanalysis}, $G_{\text{fine}}$ yields a closer representation of the original image than $G_{\text{coarse}}$ for all cases. In particular, we see that this analysis holds for high frequencies (with a \texttt{filter_rate} of 2).

\vspace{-5pt}
\subsection{Training choices}
\subsubsection{Ablation study on the perceptual losses}\label{section:Abblationperceptual}

In this discussion, we explore the importance of perceptual losses in our experiments. Table \ref{table:suptab2} illustrates that for Cityscapes and ADE20-Outdoors, the best results are achieved when the values of $\lambda_{\text{VGG}}$ and $\lambda_{\text{FM}}$ are 1 and 0.1, respectively. However, for ADE20K and COCO-stuff, larger values of $\lambda_{\text{VGG}}$ and $\lambda_{\text{FM}}$ (10 and 1, respectively) lead to the best performance. Additionally, we present an ablation study on the impact of $\lambda_{\text{VGG}}$ in Table \ref{ablation_perceptual_losses}. This study highlights the different responses of Cityscapes and ADE20K to changes in $\lambda_{\text{VGG}}$. We believe that small and specific datasets like Cityscapes and ADE20-Outdoors pose a particularly challenging problem, where the network may not easily capture the key features required to generate photo-realistic images. In such cases, perceptual losses are vital to guide learning and achieve good results. Conversely, larger datasets like COCO-stuff and ADE20K provide enough samples for the network to learn by itself, and perceptual constraints can be relaxed to promote diversity in the image generation process.

\begin{table}[t]
\centering
\scalebox{1}
{
\begin{tabular}{@{}ccccc@{}}
\toprule
\multirow{2}{*}{$\lambda_{\text{VGG}}$} & \multicolumn{2}{c}{Cityscapes} & \multicolumn{2}{c}{ADE20K} \\
& {FID $\downarrow$} & {mIoU $\uparrow$}  & {FID $\downarrow$} & {mIoU $\uparrow$} \\
\cline{1-5}
10.0&\textbf{47.10}&\textbf{70.8}&34.6&45.2\\
1.0&55.4&68.6&\textbf{28.1}&\textbf{49.0}\\
0.1&59.8&65.9&35.8&42.7\\
\bottomrule
\end{tabular}
}
\caption{\textbf{Performance of the semantic image synthesis of Pix2PixHD for different values of $\lambda_{\text{VGG}}$.} Experiments done on the second-stage training part of $G_{\text{fine}}$ only.}
\label{ablation_perceptual_losses}
\end{table}

\subsubsection{Ablation study on the data augmentation}

In this section, we investigate the impact of data augmentation for label-to-image translation on Cityscapes. We note that OASIS employs LabelMix~\cite{olsson2021classmix} data augmentation. Several mixing techniques have been proposed for tasks like semantic segmentation and semi-supervised learning, including CutMix~\cite{french2019semi} and Superpixel-mix~\cite{franchi2021robust}, which is designed to improve the robustness of semantic segmentation. We present the results of applying data augmentation during the training of Robusta, composed of $G_{\text{coarse}}$ and $G_{\text{fine}}$ only, in Table~\ref{fig:data_aug}. We find that
mixing data augmentation improves the FID but reduces the mIoU. LabelMix greatly enhances the FID,
at the cost of a significant drop in mIoU. Superpixel-mix appears to strike the best compromise. However, due to their poor performance on mIoU, we decide not to employ these strategies.

\begin{table}[t]
\centering
\scalebox{1}
{
\begin{tabular}{@{}lcc@{}}
\toprule
   Data augmentation   & FID $\downarrow$       & mIoU $\uparrow$    \\ \hline
Pix2PixHD w/ SegFormer (B5)          & 56.34 & \textbf{68.76}            \\
Robusta (B5) + CutMix   &  51.02 & 63.02               \\
Robusta (B5) + LabelMix & \textbf{44.39} & 67.25             \\
Robusta (B5) + SuperPixelMix  & 45.38 & 67.46            \\ \bottomrule
\end{tabular}
}
\caption{\textbf{Performance of Pix2PixHD and Robusta on Cityscapes} with different data augmentations.}
\label{fig:data_aug}
\end{table}

\clearpage

\section{Discussions}

\subsection{On the importance of high-quality image generation for semantic segmentation}

To evaluate the quality of images required for training a DNN, we train a Deeplab v3+ network using images generated by various cGANs. Our training protocol involves two steps and two DNNs, namely a student DNN optimized by backpropagation and a teacher DNN optimized by exponential moving average (EMA)~\cite{tarvainen2017mean}. The two training steps were conducted in parallel and involved classical supervised training on all data as well as training on images through pseudo-annotation with the teacher's DNN. This training protocol allowed the DNN to consider issues related to image synthesis.

Table \ref{table:supdatasetgene} displays our results using different image generation techniques. Our findings indicate that the Robusta-generated dataset offers the best performance when combined with real images. However, it is worth noting that using solely the images generated by Robusta did not result in ideal performance. Nevertheless, these results should be considered in relation to the performance of a model trained on GTA~\cite{richter2016playing}, which yields around 31\% of mIoU.

\begin{table}[tbh!]
\begin{center}
\scalebox{1}
{
\begin{tabular}{@{}lcc@{}}
\toprule
Dataset                            & Model    & mIoU \\ \hline
Cityscapes                         &         & 76.5 \\
Cityscapes + \textit{Corrupted-CS} & SPADE   & 76.0 \\
Cityscapes + \textit{Corrupted-CS} & OASIS   & 77.1 \\
\textit{Corrupted-CS}              & Robusta & 41.7 \\
Cityscapes + \textit{Corrupted-CS} & Robusta & \textbf{78.4} \\ \bottomrule
\end{tabular}
}
\end{center}

\caption{\textbf{Aleatoric uncertainty study} on Cityscapes-C, Foggy Cityscapes~\cite{hu2019depth} and Rainy Cityscapes~\cite{sakaridis2018semantic}.}\label{table:supdatasetgene}
\end{table}

\subsection{On semantic segmentation and uncertainties}

To achieve the goal of improving safety, it is necessary to improve our algorithms' generalization properties, which translates, for instance, into an increase of the accuracy on corrupted datasets, and enhance their anomaly detection capabilities. To do so, it is crucial to address the issue of uncertainties in semantic segmentation~\cite{kendall2017uncertainties,cygert2021closer}. In other words, to improve both generalization and anomaly detection, our algorithms must be able to handle and be robust to uncertainties in semantic segmentation.

Let us begin by introducing the uncertainties that occur in semantic segmentation~\cite{kendall2017uncertainties,cygert2021closer}. As learning algorithms, they are confronted with uncertainties, which can be categorized into two types: aleatoric and epistemic uncertainties~\cite{gawlikowski2021survey}.

Aleatoric uncertainties~\cite{hullermeier2021aleatoric} are related to the data and are caused by natural variation or lack of information. In semantic segmentation, they would mostly correspond to unclear object borders or to corruptions in the test data.

Epistemic uncertainties~\cite{hullermeier2021aleatoric} arise when the model is ill-constrained, that is that the minimization of the loss on the training set does not properly define the outputs. This would correspond to unknown shapes or textures on the inputs involving misspecified outputs.

\subsection{Training speed}\label{section:velocity}

\begin{figure}[t]
\centering
\begin{center}
\includegraphics[width=0.9\linewidth]{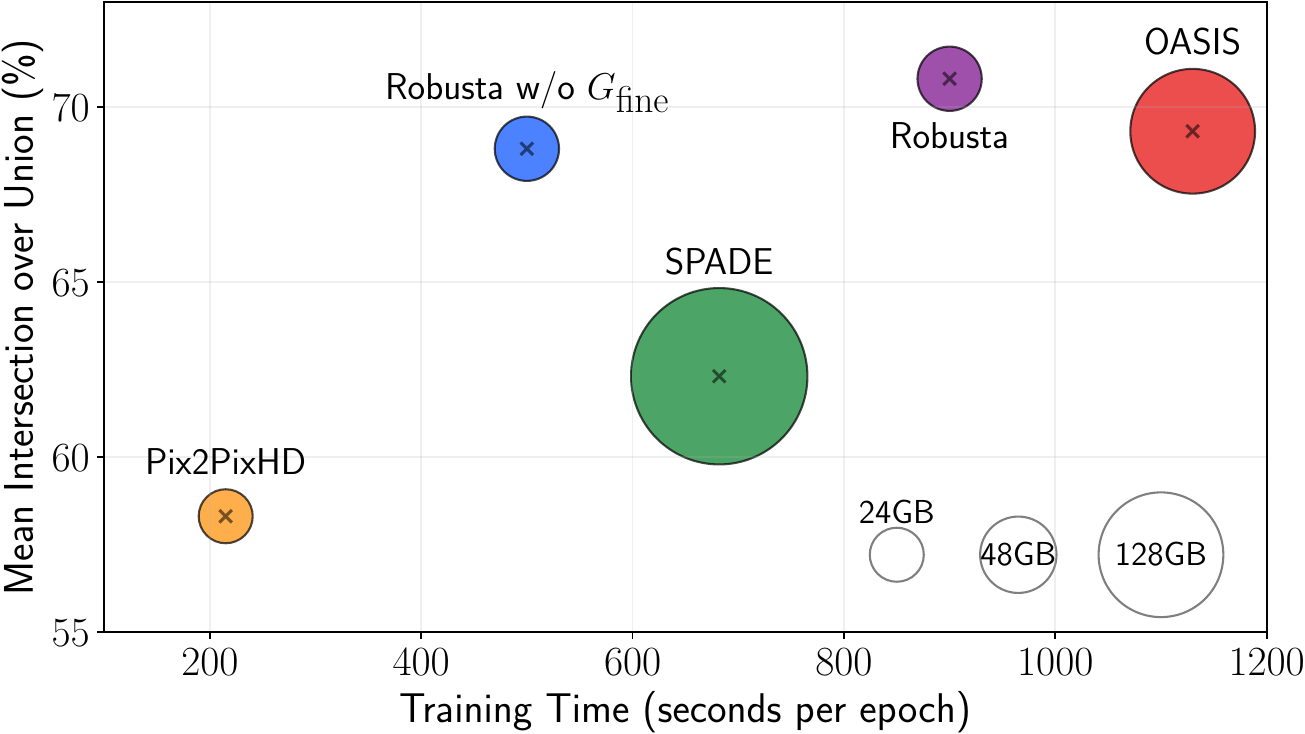}
\end{center}
\caption{\textbf{Computation cost and performance trade-offs for several image-to-image translation techniques on Cityscapes.} The y-axis shows the FID, and the x-axis shows the training time for one epoch with a batch size of 4. Training time is averaged over 5 epochs. The circle area is proportional to the required VRAM in GB to train the model. In the case of Robusta, $G_{\text{coarse}}$ and $G_{\text{fine}}$ are trained sequentially, and the required memory corresponds to the heaviest generator, $G_{\text{coarse}}$. The best approaches are closer to the upper-left corner.}
\label{fig:Computationcost}
\end{figure}

Our aim is for the algorithm to not only produce realistic images and perform well in terms of robustness but also to be scalable and {require} a reasonable amount of computational resources that match the complexity of the task at hand. To further explain this point, we analyze the balance between the training time and the FID score of the latest and most competitive algorithms for image-to-label translation. The results, as depicted in Figure \ref{fig:Computationcost}, show that Robusta is positioned in a favorable spot. Compared to Pix2PixHD, it has fewer parameters, faster training time than OASIS, and a better FID score than other algorithms. By incorporating \(G_{\text{fine}}\), the training time is slightly affected, but there is a significant performance improvement. It's important to note that the reason for OASIS's low training time is due to the per-pixel discriminator that predicts full-resolution segmentation maps, which requires a large amount of VRAM, as shown in Table \ref{tab:complexity}. Additionally, the LabelMix data-augmentation strategy utilized by the algorithm is not as efficient regarding GPU usage.

Table \ref{tab:complexity} provides additional information on other relevant metrics about scalability. It should be noted that the number of parameters and VRAM is more closely related to space complexity. Since our training process involves two stages, the effective number of parameters/VRAM experienced during training corresponds to the values of the first stage, which is a bottleneck due to the lighter weight of \(G_{\text{fine}}\) compared to \(G_{\text{coarse}}\). Overall, Robusta is comparable to SPADE and OASIS regarding the number of parameters and Mult-Adds. However, our algorithm utilizes only half as much VRAM as OASIS.

\begin{table}[t]
\centering
\scalebox{0.9}
{
\begin{tabular}{@{}lccc@{}}
\toprule
Architecture & MA $\downarrow$ & Params $\downarrow$ & VRAM $\downarrow$ \\
\midrule
Pix2PixHD & 161.4 & 188.1  & 24  \\
SPADE & 272.5 & 98.6  & 256 \\
OASIS & 296.8 & 93.4  & 128 \\
Robusta w/o $G_{\text{fine}}$  & 393.6 & 98.6 & 34 \\
$G_{\text{fine}}$  & 297.3 & 50.5 & 24 \\
\bottomrule
\end{tabular}
}
\caption{\textbf{MA: Mult-Adds} corresponds to the number of Giga multiply-add operations of a forward pass through both the generators and discriminators with a single input. \textbf{Params: total number of parameters (in million)} of the generators and discriminators combined. \textbf{VRAM:} memory (in GB) required for a forward/backward through both the generators and discriminators. \textbf{MA} and \textbf{Params} are estimated with Torchinfo~\cite{Torchinfo}.}
\label{tab:complexity}
\end{table}

\subsection{Evaluation protocol}

Comparing the results of generative networks is often challenging, and thus, either the FID or mIoU is commonly used. While human assessment is also used for qualitative validation, it can be time-consuming, and evaluators may introduce biases.

We note that the mIoU could also have a bias. Indeed, we see that most of the techniques outperform the network mIoU trained on the training set and tested on the real validation set. More specifically, the mIoU of DRN~\cite{yu2017dilated} on Cityscapes equals $66.35\%$, the one of Deeplabv2~\cite{chen2017deeplab} on COCO-stuff is $39.03\%$, and the one of UperNet101~\cite{xiao2018unified} on ADE20K is $42.74\%$. Hence, the approaches that beat these values can transform the validation set into a training set.

In their work on domain adaptation, Ben David et al. \cite{ben2010theory} mention the divergence between the distribution of the target domain and that of the source domain, which would limit the empirical risk for the target domain. Modern approaches seem to be finding ways to facilitate the transfer between the source and target domains. One small concern is that access to the label is necessary, which is still a problem. However, replacing it with a depth map and generating RGB images from that depth map should be possible.

\section{Qualitative results}\label{sec:qualitative}
\begin{figure*}[t]
\centering
\begin{tabular}{llll}
\rotr{\begin{tabular}[c]{@{}c@{}}\scriptsize{Label}\end{tabular}} & \includegraphics[width=0.30\linewidth]{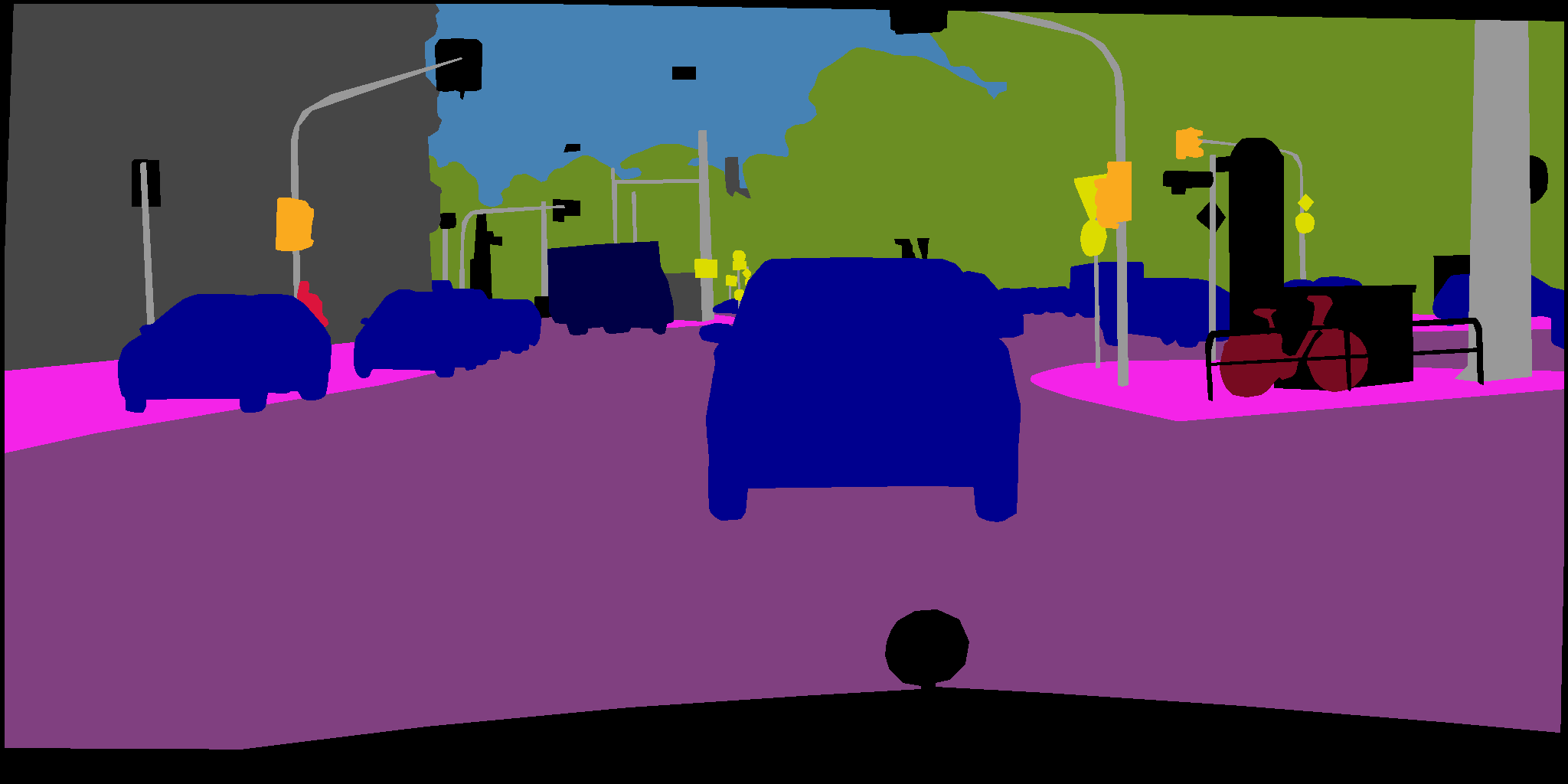}& \includegraphics[width=0.30\linewidth]{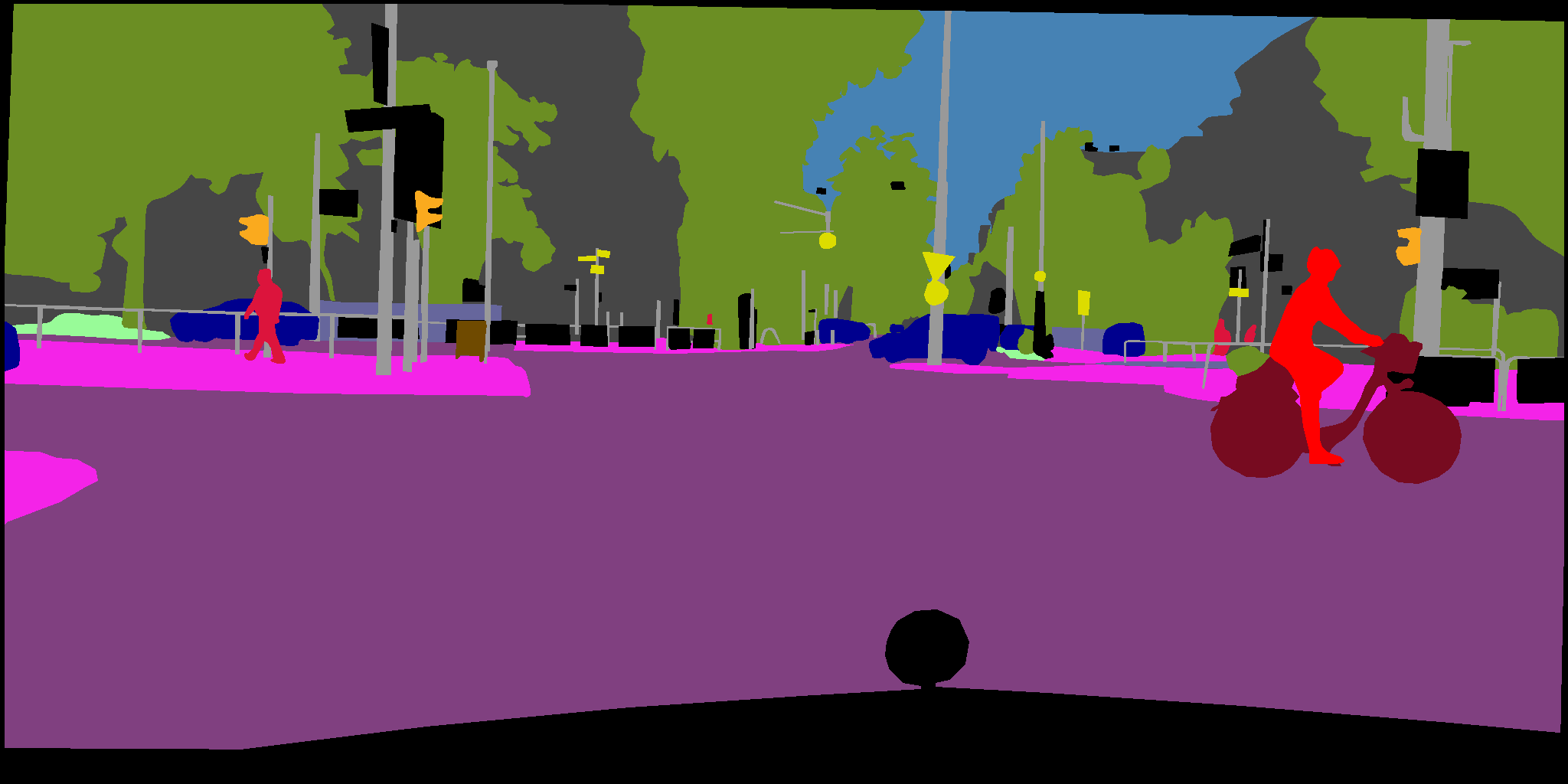} & \includegraphics[width=0.30\linewidth]{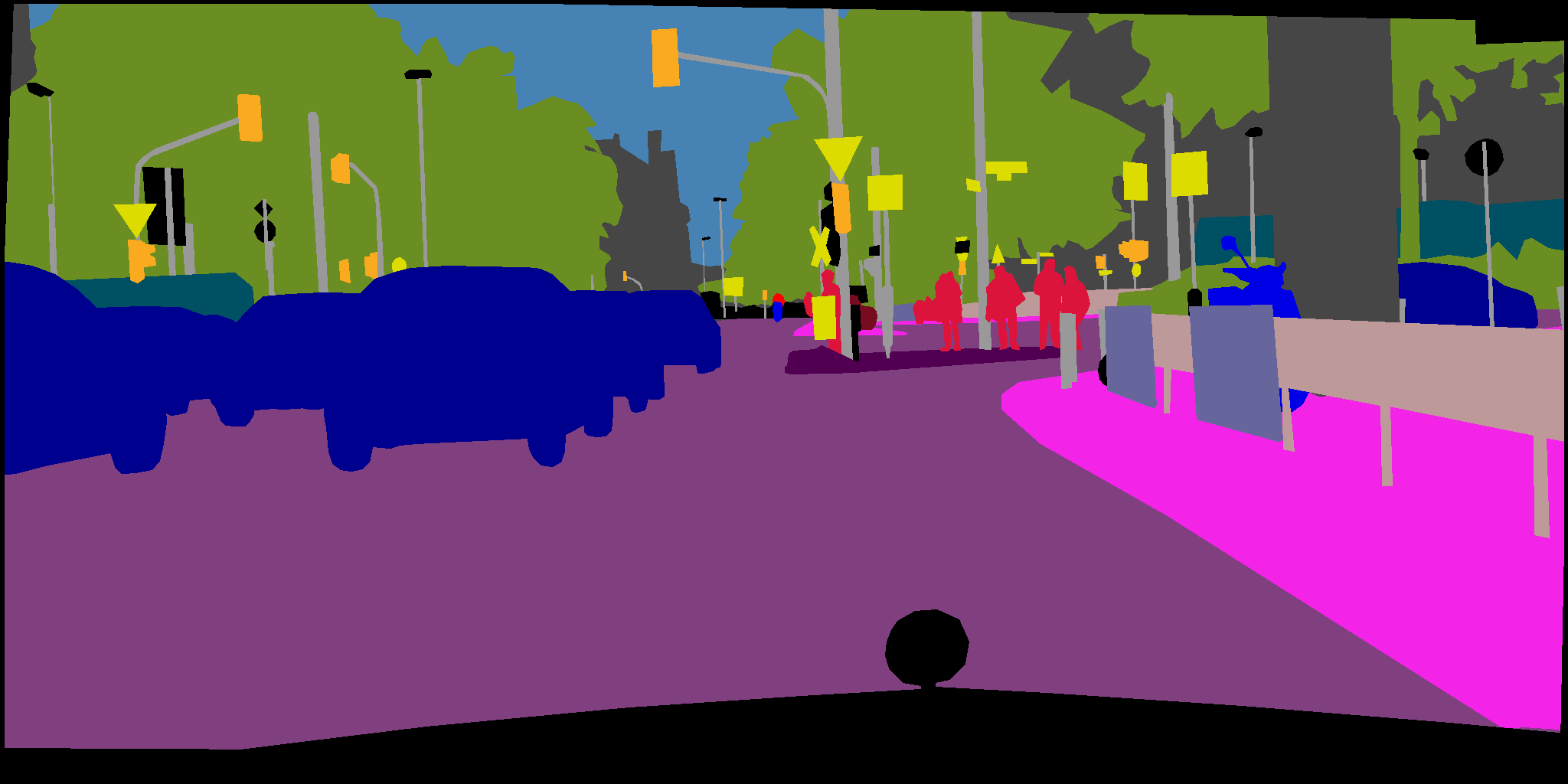} \\
\rotr{\begin{tabular}[c]{@{}c@{}}\scriptsize{Image}\end{tabular}} & \includegraphics[width=0.30\linewidth]{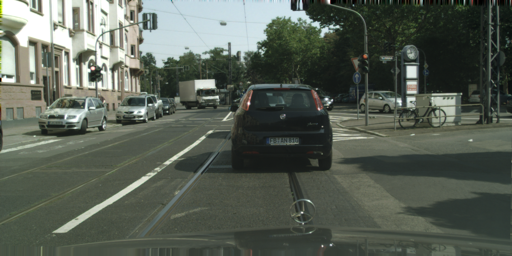} & \includegraphics[width=0.30\linewidth]{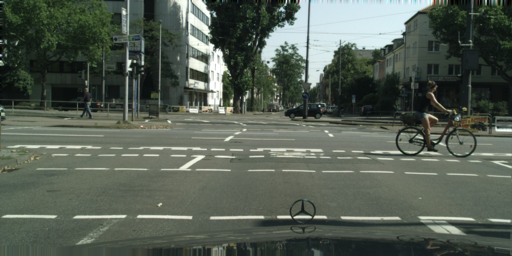} & \includegraphics[width=0.30\linewidth]{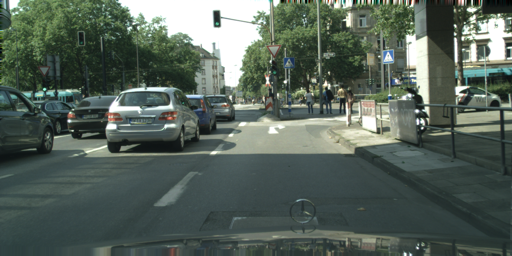} \\
\rotr{\begin{tabular}[c]{@{}c@{}}\scriptsize{Oasis}\\\scriptsize{prediction}\end{tabular}} & \includegraphics[width=0.30\linewidth]{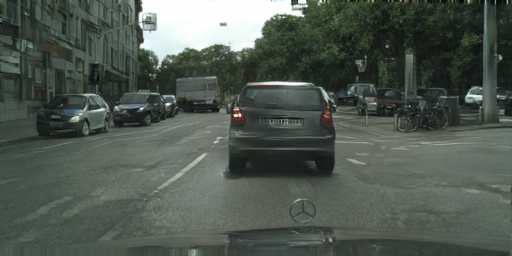} & \includegraphics[width=0.30\linewidth]{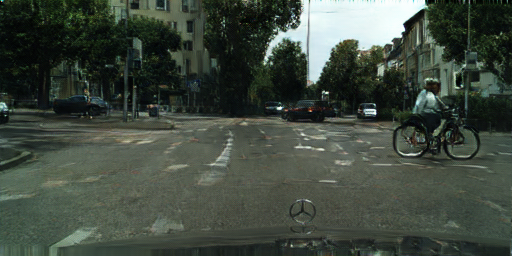} & \includegraphics[width=0.30\linewidth]{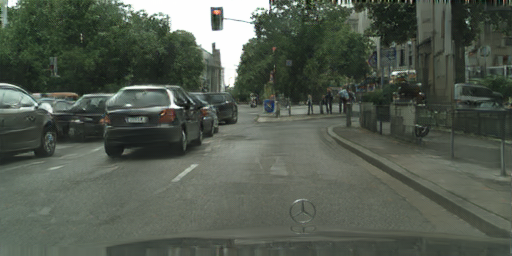} \\
\rotr{\begin{tabular}[c]{@{}c@{}}\scriptsize{Robusta}\\\scriptsize{prediction}\end{tabular}} & \includegraphics[width=0.30\linewidth]{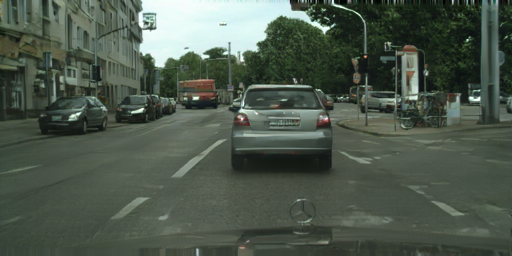}  & \includegraphics[width=0.30\linewidth]{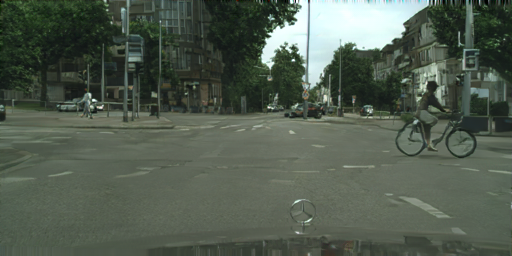}  & \includegraphics[width=0.30\linewidth]{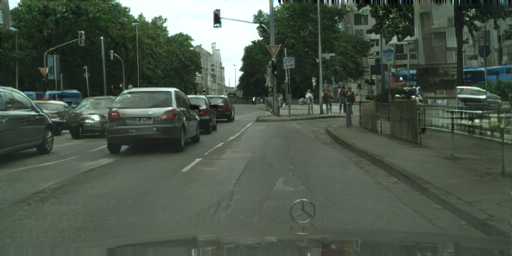}
\end{tabular}
\caption{Qualitative comparison of Robusta with other methods on Cityscapes.}
\label{fig:results-Cityscape}
\end{figure*}

\begin{figure*}[t]
\centering
\begin{tabular}{lllll}
\rotr{\begin{tabular}[c]{@{}c@{}}\scriptsize{Label}\end{tabular}} & \includegraphics[width=0.22\linewidth]{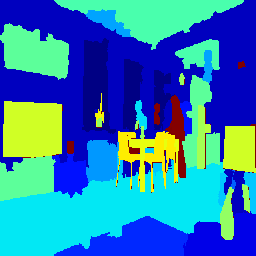}& \includegraphics[width=0.22\linewidth]{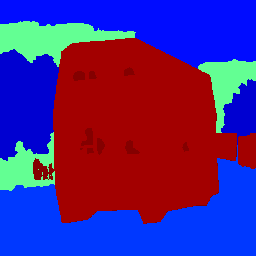} &
\includegraphics[width=0.22\linewidth]{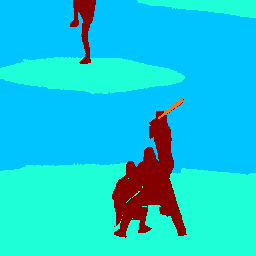} &
\includegraphics[width=0.22\linewidth]{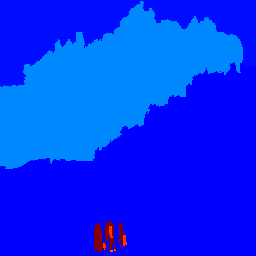} \\
\rotr{\begin{tabular}[c]{@{}c@{}}\scriptsize{Image}\end{tabular}} & \includegraphics[width=0.22\linewidth]{images/supplementary/000000000139_gt.png}& \includegraphics[width=0.22\linewidth]{images/supplementary/000000001584_gt.png} &
\includegraphics[width=0.22\linewidth]{images/supplementary/000000002153_gt.png} &
\includegraphics[width=0.22\linewidth]{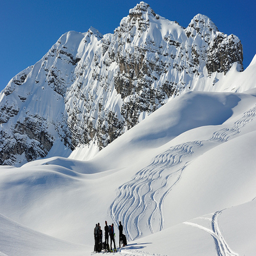} \\
\rotr{\begin{tabular}[c]{@{}c@{}}\scriptsize{Oasis}\\\scriptsize{prediction}\end{tabular}} & \includegraphics[width=0.22\linewidth]
{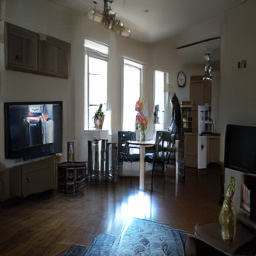} & \includegraphics[width=0.22\linewidth]{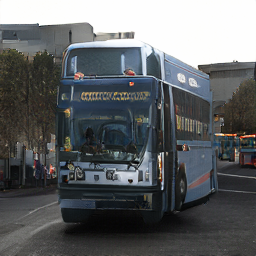} &
\includegraphics[width=0.22\linewidth]{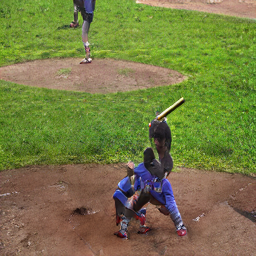} &
\includegraphics[width=0.22\linewidth]{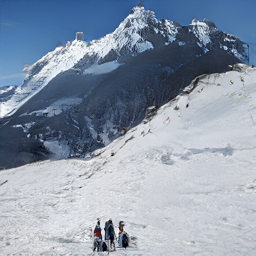} \\
\rotr{\begin{tabular}[c]{@{}c@{}}\scriptsize{Robusta}\\\scriptsize{prediction}\end{tabular}} & \includegraphics[width=0.22\linewidth]{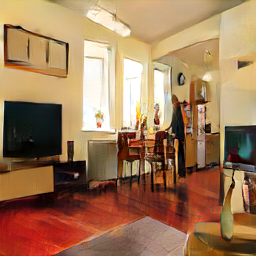}  & \includegraphics[width=0.22\linewidth]{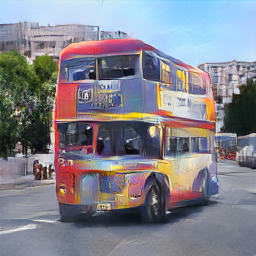}  & \includegraphics[width=0.22\linewidth]{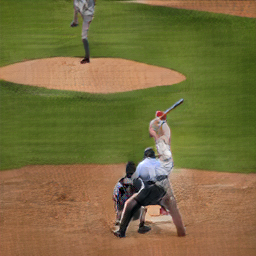}  & \includegraphics[width=0.22\linewidth]{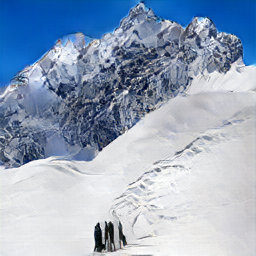}
\end{tabular}
\caption{Qualitative comparison of Robusta with other methods on COCO-stuff.
}
\label{fig:results-COCO}
\end{figure*}

\begin{figure*}[t]
\centering
\begin{tabular}{lllll}
\rotr{\begin{tabular}[c]{@{}c@{}}\scriptsize{Label}\end{tabular}} & \includegraphics[width=0.22\linewidth]{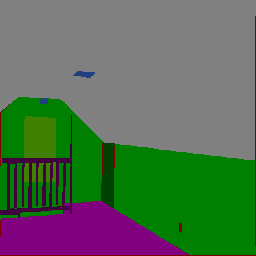}& \includegraphics[width=0.22\linewidth]{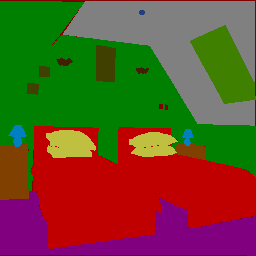} &
\includegraphics[width=0.22\linewidth]{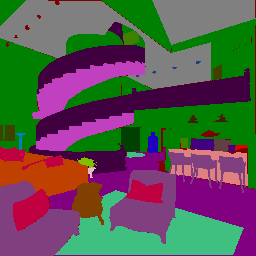} &
\includegraphics[width=0.22\linewidth]{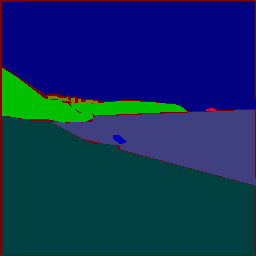} \\
\rotr{\begin{tabular}[c]{@{}c@{}}\scriptsize{Image}\end{tabular}} & \includegraphics[width=0.22\linewidth]{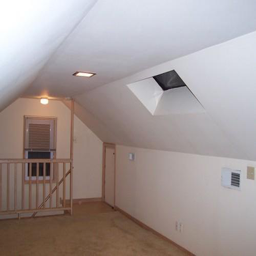}& \includegraphics[width=0.22\linewidth]{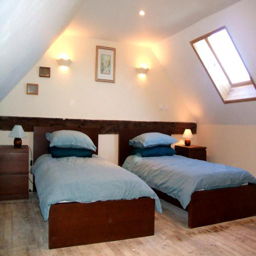} &
\includegraphics[width=0.22\linewidth]{images/supplementary/ade20k/gt_526.png} &
\includegraphics[width=0.22\linewidth]{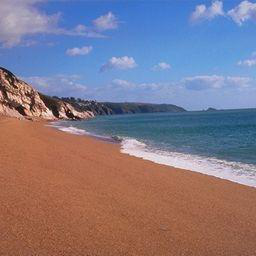} \\
\rotr{\begin{tabular}[c]{@{}c@{}}\scriptsize{Oasis}\\\scriptsize{prediction}\end{tabular}} & \includegraphics[width=0.22\linewidth]
{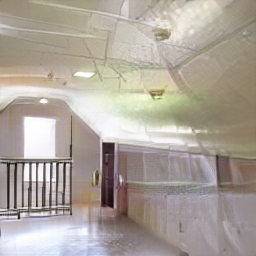} & \includegraphics[width=0.22\linewidth]{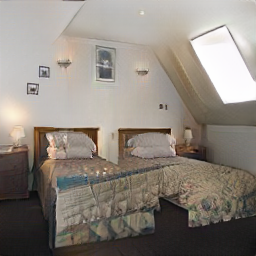} &
\includegraphics[width=0.22\linewidth]{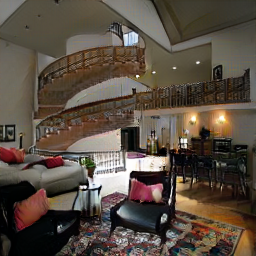} &
\includegraphics[width=0.22\linewidth]{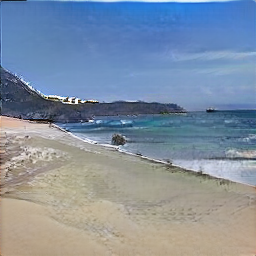} \\
\rotr{\begin{tabular}[c]{@{}c@{}}\scriptsize{Robusta}\\\scriptsize{prediction}\end{tabular}} & \includegraphics[width=0.22\linewidth]{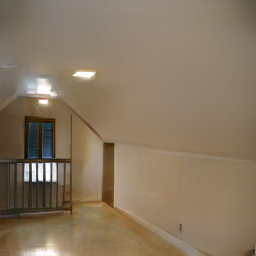}  & \includegraphics[width=0.22\linewidth]{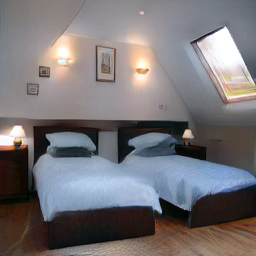}  & \includegraphics[width=0.22\linewidth]{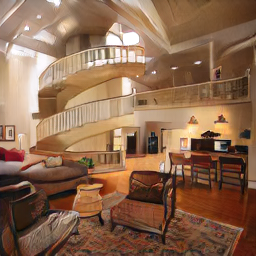}  & \includegraphics[width=0.22\linewidth]{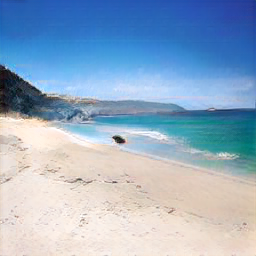}
\end{tabular}
\caption{Qualitative comparison of Robusta with other methods on ADE20K.
}
\label{fig:results-ADE20K}
\end{figure*}

\begin{figure*}[t]
\centering
\begin{tabular}{llll}
\rotr{\begin{tabular}[c]{@{}c@{}}\scriptsize{Label}\end{tabular}} & \includegraphics[width=0.30\linewidth]{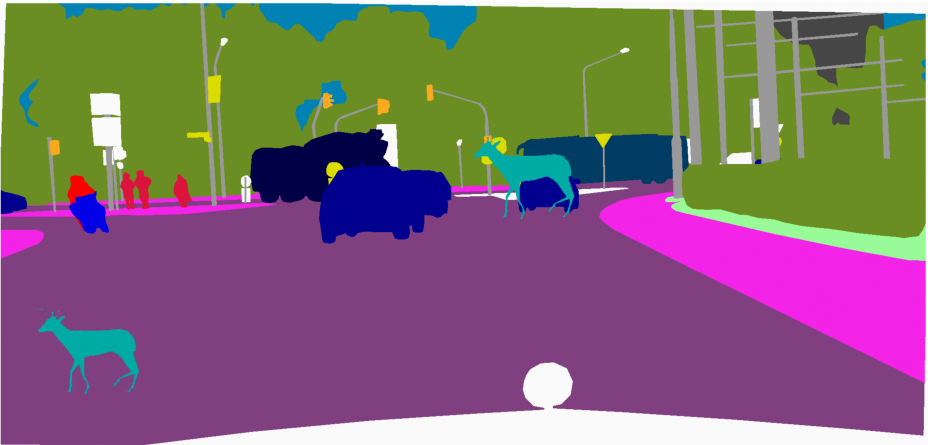}& \includegraphics[width=0.30\linewidth]{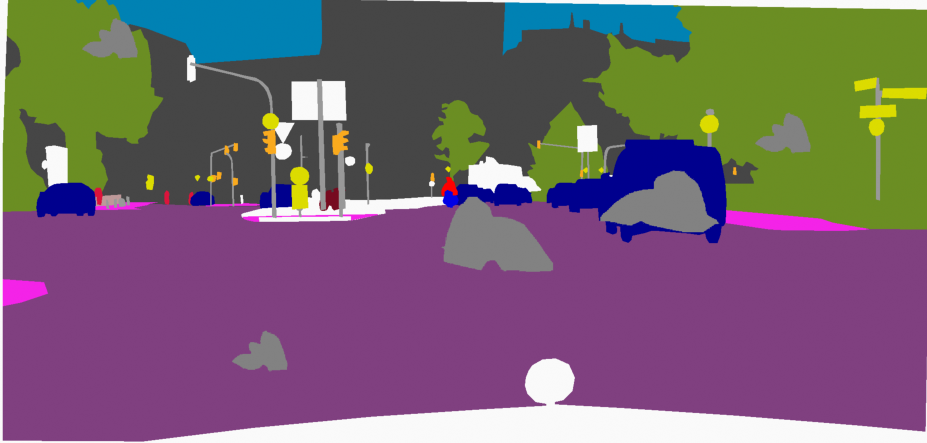} & \includegraphics[width=0.30\linewidth]{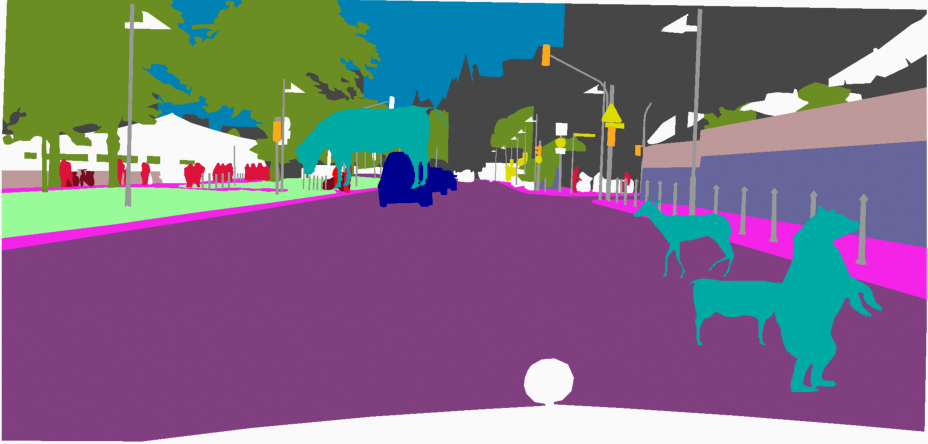} \\
\rotr{\begin{tabular}[c]{@{}c@{}}\scriptsize{SPADE}\\\scriptsize{prediction}\end{tabular}} & \includegraphics[width=0.30\linewidth]{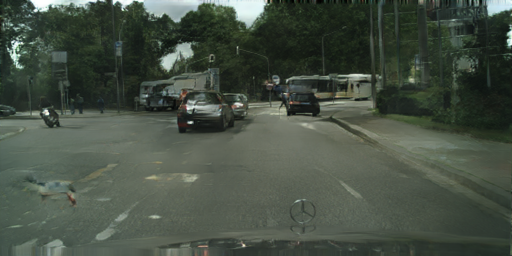} & \includegraphics[width=0.30\linewidth]{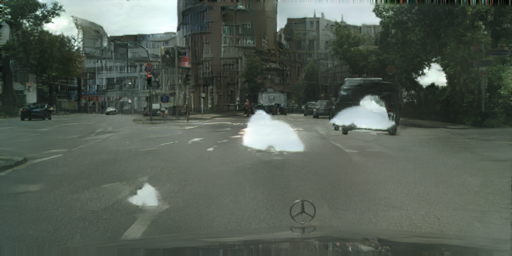} & \includegraphics[width=0.30\linewidth]{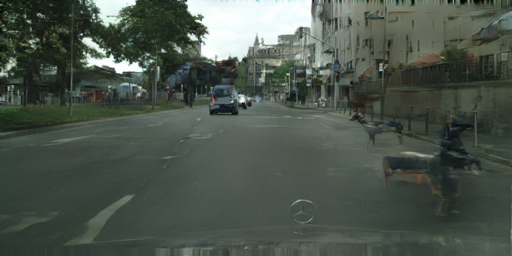} \\
\rotr{\begin{tabular}[c]{@{}c@{}}\scriptsize{Oasis}\\\scriptsize{prediction}\end{tabular}} & \includegraphics[width=0.30\linewidth]{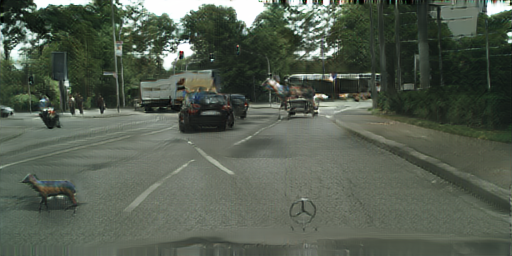} & \includegraphics[width=0.30\linewidth]{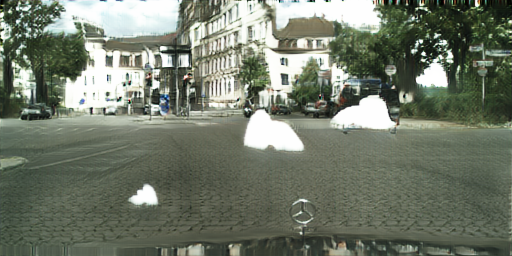} & \includegraphics[width=0.30\linewidth]{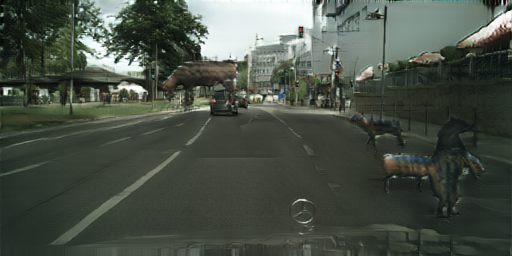} \\
\rotr{\begin{tabular}[c]{@{}c@{}}\scriptsize{Robusta}\\\scriptsize{prediction}\end{tabular}} & \includegraphics[width=0.30\linewidth]{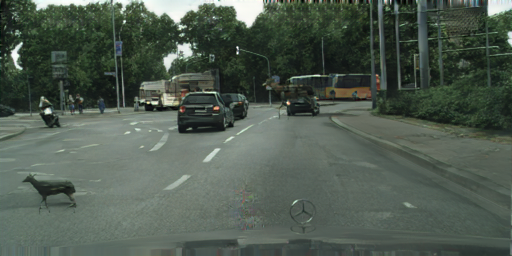}  & \includegraphics[width=0.30\linewidth]{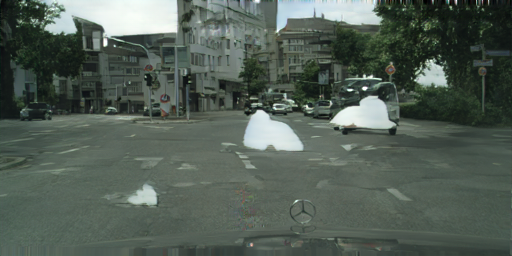}  & \includegraphics[width=0.30\linewidth]{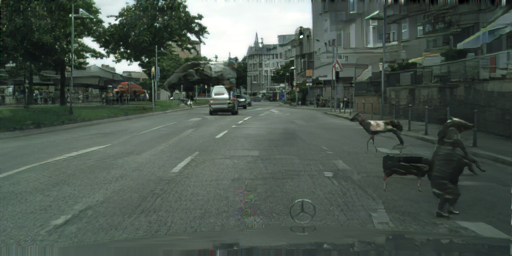}

\end{tabular}
\caption{Qualitative comparison of Robusta with other methods on \emph{Outlier-Cityscapes}.
}
\label{fig:results-Outlier-Cityscape}
\end{figure*}

\begin{figure*}[t]
\centering
\begin{tabular}{llll}
\rotr{\begin{tabular}[c]{@{}c@{}}\scriptsize{Label}\end{tabular}} & \includegraphics[width=0.30\linewidth]{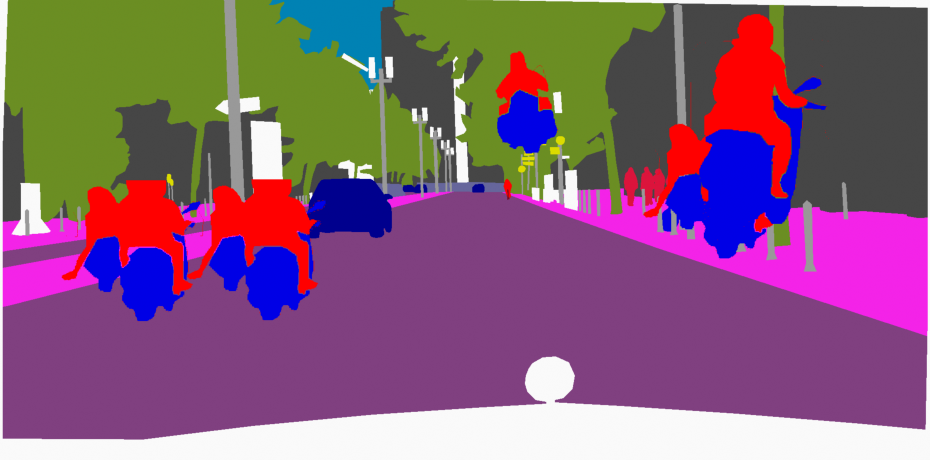}& \includegraphics[width=0.30\linewidth]{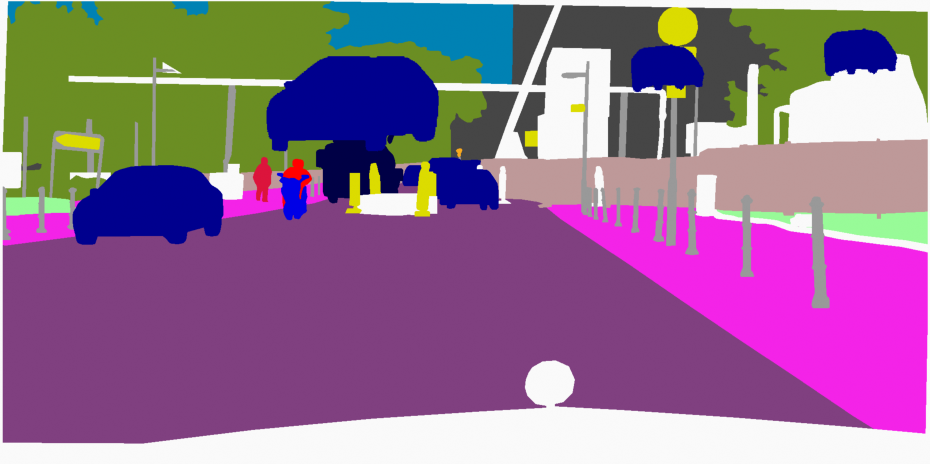} & \includegraphics[width=0.30\linewidth]{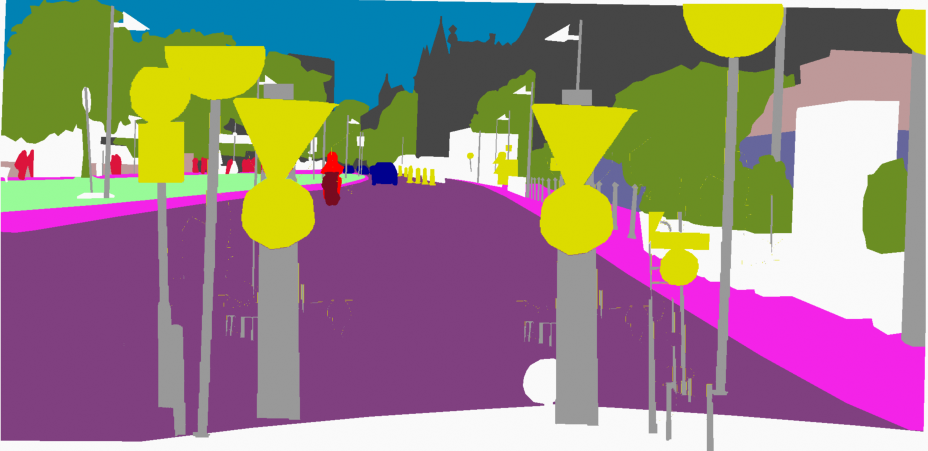} \\
\rotr{\begin{tabular}[c]{@{}c@{}}\scriptsize{SPADE}\\\scriptsize{prediction}\end{tabular}} & \includegraphics[width=0.30\linewidth]{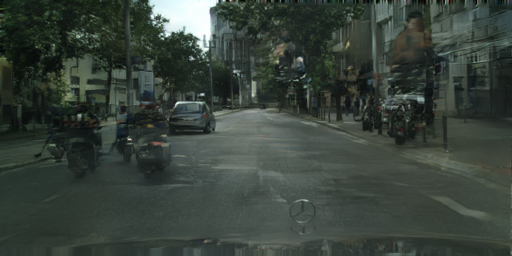} & \includegraphics[width=0.30\linewidth]{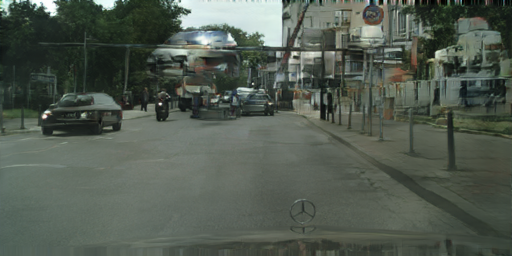} & \includegraphics[width=0.30\linewidth]{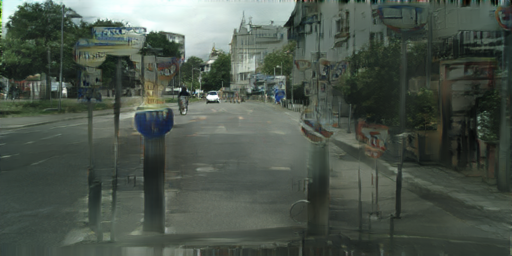} \\
\rotr{\begin{tabular}[c]{@{}c@{}}\scriptsize{Oasis}\\\scriptsize{prediction}\end{tabular}} & \includegraphics[width=0.30\linewidth]{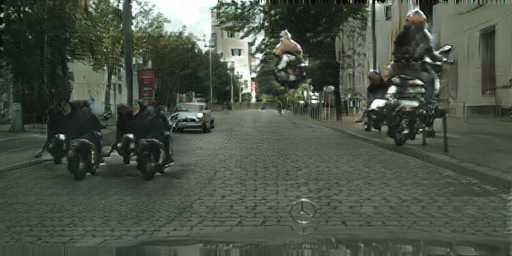} & \includegraphics[width=0.30\linewidth]{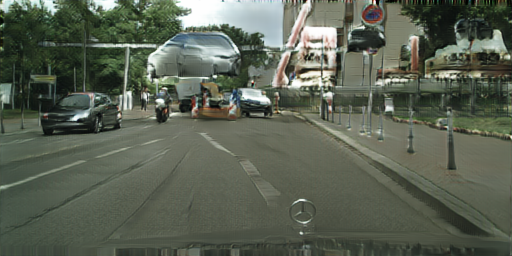} & \includegraphics[width=0.30\linewidth]{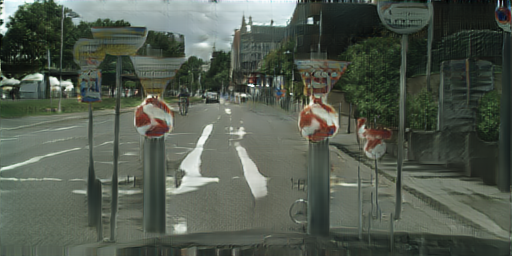} \\
\rotr{\begin{tabular}[c]{@{}c@{}}\scriptsize{Robusta}\\\scriptsize{prediction}\end{tabular}} & \includegraphics[width=0.30\linewidth]{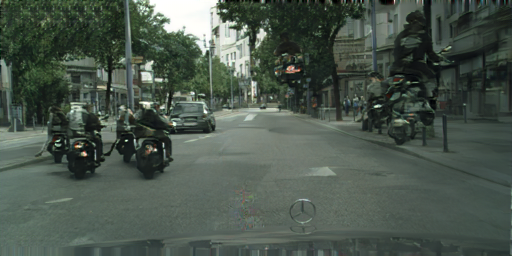}  & \includegraphics[width=0.30\linewidth]{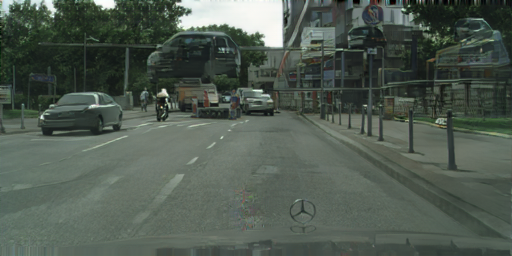}  & \includegraphics[width=0.30\linewidth]{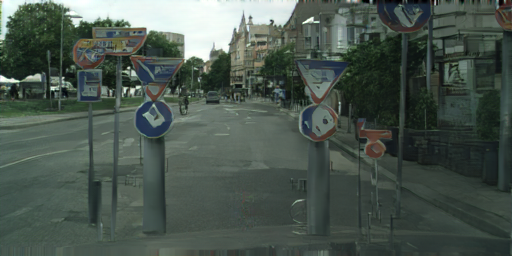}

\end{tabular}
\caption{Qualitative comparison of Robusta with other methods on \emph{Corrupted-Cityscapes}.
}
\label{fig:results-corrupted-Cityscape}
\end{figure*}

\begin{figure*}[t]
\centering
\begin{tabular}{llll}
\rotr{\begin{tabular}[c]{@{}c@{}}\scriptsize{Infrared image}\end{tabular}} & \includegraphics[width=0.30\linewidth]{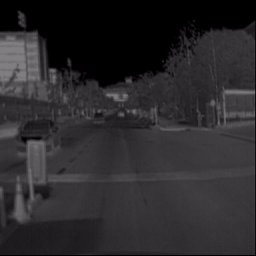}& \includegraphics[width=0.30\linewidth]{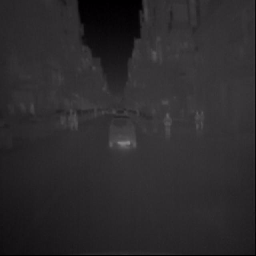}&
\includegraphics[width=0.30\linewidth]{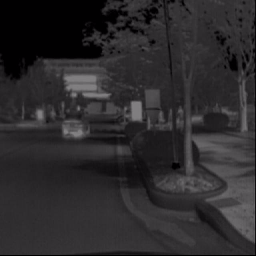}\\
\rotr{\begin{tabular}[c]{@{}c@{}}\scriptsize{Visible image}\end{tabular}} & \includegraphics[width=0.30\linewidth]{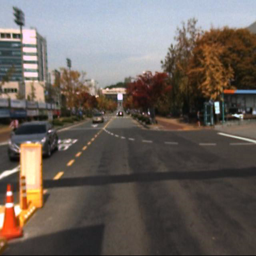} &
\includegraphics[width=0.30\linewidth]{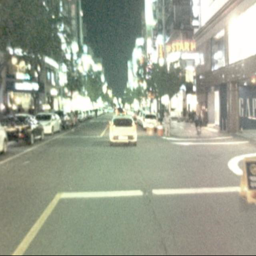}&
\includegraphics[width=0.30\linewidth]{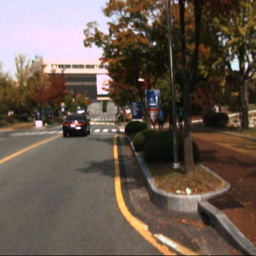}\\
\rotr{\begin{tabular}[c]{@{}c@{}}\scriptsize{Pix2PixHD}\\\scriptsize{prediction}\end{tabular}} & \includegraphics[width=0.30\linewidth]
{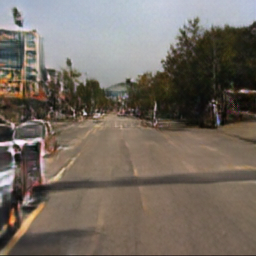} &
\includegraphics[width=0.30\linewidth]{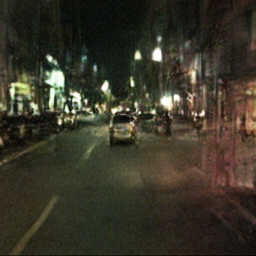}&
\includegraphics[width=0.30\linewidth]{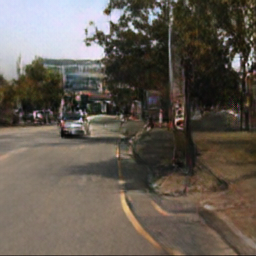}\\
\rotr{\begin{tabular}[c]{@{}c@{}}\scriptsize{Robusta}\\\scriptsize{prediction}\end{tabular}} & \includegraphics[width=0.30\linewidth]{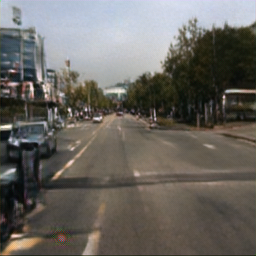}  & \includegraphics[width=0.30\linewidth]{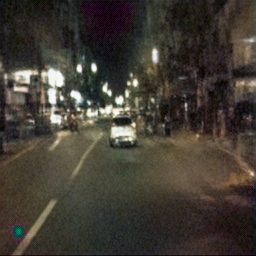} &
\includegraphics[width=0.30\linewidth]{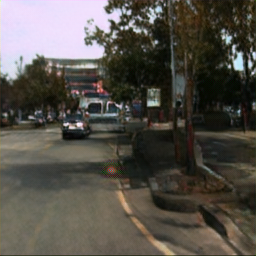}
\end{tabular}
\caption{Qualitative comparison of Robusta with other methods on KAIST.
}
\label{fig:results-KAIST}
\end{figure*}

In this section, we illustrate the comparison between OASIS and Robusta with some visual examples. Experiments are done on three datasets: Cityscapes~\cite{cordts2016cityscapes} in Figure \ref{fig:results-Cityscape}, COCO-stuff~\cite{lin2014microsoft} in \ref{fig:results-COCO}, ADE20K~\cite{zhou2017scene} in \ref{fig:results-ADE20K}, \emph{Outlier-Cityscapes} in \ref{fig:results-Outlier-Cityscape}, \emph{Corrupted-Cityscapes} in \ref{fig:results-corrupted-Cityscape}, and KAIST~\cite{jeong2019kaist} in Figure \ref{fig:results-KAIST}.

\end{document}